\documentclass{article}

\usepackage{nac_preprint}

\usepackage[utf8]{inputenc} 
\usepackage[T1]{fontenc}    
\usepackage{hyperref}       
\usepackage{url}            
\usepackage{booktabs}       
\usepackage{amsfonts}       
\usepackage{nicefrac}       
\usepackage{microtype}      

\usepackage[utf8]{inputenc} 
\usepackage[T1]{fontenc}    
\usepackage{hyperref}       
\usepackage{url}            
\usepackage{booktabs}       
\usepackage{amsfonts}       
\usepackage{nicefrac}       
\usepackage{microtype}      
\usepackage{lipsum}
\usepackage{wrapfig}

\usepackage{booktabs} 
\usepackage{subfig}
\usepackage{amsmath,amssymb}
\usepackage{mathtools}
\usepackage[toc,page]{appendix}
\usepackage{enumitem}
\usepackage{graphics}
\usepackage{graphicx}
\usepackage{algorithm}
\usepackage{algorithmicx}
\usepackage[noend]{algpseudocode}
\algnewcommand{\LineComment}[1]{\State \(//\) #1}
\usepackage{array}

\title{Large-Scale Gradient-Free Deep Learning with Recursive Local Representation Alignment}

\author{%
  Alexander Ororbia* \\
  Rochester Institute of Technology \\
  Rochester, NY 14623 \\
  \texttt{ago@cs.rit.edu} \\
  \And
  Ankur Mali* \\
  The Pennsylvania State University \\
  State College, PA 16801 \\
  \texttt{aam35@psu.edu} \\
  \AND 
  Daniel Kifer \\
  The Pennsylvania State University \\
  State College, PA 16801 \\
  \texttt{duk17@psu.edu} \\
  \And 
  C. Lee Giles \\
  The Pennsylvania State University \\
  State College, PA 16801 \\
  \texttt{clg20@psu.edu} \\
}

\begin{document}

\setlength{\abovedisplayskip}{3pt}
\setlength{\belowdisplayskip}{3.1pt}

\maketitle

\begin{abstract}
Training deep neural networks on large-scale datasets requires significant hardware resources whose costs (even on cloud platforms) put them out of reach of smaller organizations, groups, and individuals. Backpropagation, the workhorse for training these networks, is an inherently sequential process that is difficult to parallelize. Furthermore, it requires researchers to continually develop various tricks, such as specialized weight initializations and activation functions, in order to ensure a stable parameter optimization. Our goal is to seek an effective, neuro-biologically-plausible alternative to backprop that can be used to train deep networks. In this paper, we propose a gradient-free learning procedure, \emph{recursive local representation alignment}, for training large-scale neural architectures. 
Experiments with residual networks on CIFAR-10 and the large benchmark, ImageNet, show that our algorithm generalizes as well as backprop while converging sooner due to weight updates that are parallelizable and computationally less demanding. This is empirical evidence that a backprop-free algorithm can scale up to larger datasets. 
\end{abstract}

\section{Introduction} 
\label{sec:intro}
At the heart of training artificial neural networks (ANNs) is the calculation of adjustments that need to be made to parameters given some data. This calculation is used in tandem with an optimization procedure, such as a stochastic hill climbing procedure, to then alter the ANN's actual parameters in order to ensure it makes better future predictions. This adjustment process entails using an algorithm that can conduct credit assignment, i.e., the task of determining the contribution that individual neuronal units (within the ANN) make to the system's overall error. To conduct credit assignment and compute weight updates in state-of-the-art networks today, backprop \cite{rumelhart1986learning} is the popular algorithm of choice but has been long criticized as neuro-biologically implausible \cite{crick1989recent}. 
While backprop provides a theoretical basis for training networks, i.e. gradient descent, it also presents practical challenges, e.g., exploding/vanishing gradients \cite{glorot2010understanding}. 

In order to deal with the problems posed by backprop, researchers must resort to tricks and heuristics, e.g., careful initialization of weights, often following from a network-specific analysis of backprop's learning dynamics \cite{glorot2010understanding,he2015delving,Sussillo14,MishkinM15} or modifying network structure, for example by using ReLU instead of sigmoid activations. 
Challenges such as these prevent new users from exploiting the benefits of deep learning in novel applications (that have no pre-trained models)
and divert attention from designing models that can solve defined problems.
Furthermore, backprop is sequential in nature -- layers are updated in order, reducing opportunities for parallelization.
This limits how well we can exploit the processing power afforded by multi-CPU/GPU setups.

This paper seeks to demonstrate that a biologically-motivated algorithm can scale up to the training of large-scale architectures for large databases. Specifically, we will present a procedure that is better suited to parallelization, adjusting synaptic weight parameters with rules that are local in nature (in particular, layers can be updated out of order).
The contributions of this work are as follows:
(1) The algorithm, \emph{recursive local representation alignment} (rec-LRA), is proposed for training large-scale ANNs. Results show that it handles non-differentiable activations, converges faster than backprop, and offers faster training for large-scale benchmarks (ImageNet), and (2), Strong generalization across several datasets, including the benchmark ImageNet, is demonstrated for architectures trained using rec-LRA. Furthermore, we significantly reduce total parameter count of our networks by utilizing fast, fixed noise maps in place of convolution.

\section{Related Work}
\label{sec:related_work}

It has long since been a desire of connectionist researchers to develop learning algorithms that simultaneously are biologically-plausible and yield robust generalization to out-of-sample patterns \cite{hinton2002training,chalasani2013deep,scellier2017equilibrium,goyal2017variational,sacramento2018dendritic,nokland2019training,krotov2019unsupervised}.
One key motivation behind the development of alternative algorithms is the removal of the required symmetry between forward pathways for inference and backwards pathways for credit assignment, as is required by backprop. This has also been referred to as the weight-transport problem \cite{grossberg_resonance_1987,liao2016important}, a strong neuro-biological criticism of backprop as well as one source of its practical issues. Algorithms such as random feedback alignment (FA) \cite{lillicrap2016random} and direct feedback alignment (DFA) \cite{nokland2016direct} have shown that learning is possible, surprisingly, even if the feedback pathway is partially decoupled and random, fixed weights are used to transmit derivative signals backward. FA simply replaces the transpose of the feedforward weights in backprop with a similarly-shaped random matrix while DFA directly wires the output layer's pre-activation derivative to each layer's post-activation -- both algorithms use these random matrices to generate proxies for the partial derivatives normally given by backprop. Under a proposed framework known as discrepancy reduction, it has been shown in \cite{ororbia2017learning}
that these feedback loops are better suited for generating target representations, entirely removing the global feedback pathway of backprop -- a key idea our algorithm builds on. Algorithms such as target propagation \cite{lee2015targetprop,bartunov2018assessing,ahmad2020gait}, which are also subsumed by the discrepancy reduction framework, 
generate targets through an auto-encoding framework (a decoder attempts to approximate the inverse of a forward encoder's underlying function).

The idea of local learning, with origins in the classical frameworks of Hebbian \cite{hebb1949organization}, anti-Hebbian \cite{foldiak1990forming}, and competitive learning \cite{rumelhart1985feature}, has slowly begun to gain increased attention in the training of ANNs. Recent proposals have included decoupled neural interfaces \cite{jaderberg2016decoupled}, greedy relaxations of backprop \cite{belilovsky2019decoupled}, and others \cite{balduzzi2015kickback,taylor2016training}. Furthermore, \cite{xie2003equivalence} demonstrated that neural models using simple local Hebbian updates (in a predictive coding framework) could efficiently conduct supervised learning. Earlier approaches that employed local learning included the layer-wise training procedures that were once used to pre-train networks \cite{vincent2008extracting,bengio_greedy_2007,lee_deeply-supervised_2014,ororbia_deep_hybrid_2015a}.  The problem with these older approaches is that they were greedy--a model was built from the bottom-up, freezing lower-level parameters as higher-level feature detectors were learned. However, modern, improved generalizations have been proposed \cite{belilovsky2018greedy}.

\section{Recursive Local Representation Alignment}
\label{sec:rec_lra}
In this section, we first define our problem and present notation and then present our proposed algorithm in detail.

\subsection{The Problem \& Notation}
\label{sec:definitions_notation}
While our algorithm could be applied to any type of neural architecture (including recurrent ones), in this paper, we will focus on ones that attempt to learn a nonlinear mapping $f_\Theta$ from inputs $\mathbf{x}$ to outputs $\mathbf{y}$.
As usual, each input example can be modeled as a matrix  $\mathbf{x} \in \mathcal{R}^{I \times C}$ (e.g., for images with $I$ pixels and $C$ channels) or vector $\mathbf{x} \in \mathcal{R}^{I}$  (e.g.,  for grey-scale images with $I$ pixels or text document vectors with $I$ distinct tokens),\footnote{Vectors and matrices are assumed to be in column-major form.} or even as tensors.
On the other hand, the target $\mathbf{y} \in \mathcal{R}^{Y}$ can be modeled as a one-hot encoding, where $Y$ is the number of distinct classes/categories in a dataset.

The nonlinear mapping $f_\Theta(\mathbf{x})$ contains a set of learnable parameters housed in the construct $\Theta$, which are what algorithms such as backprop are trying to modify to improve predictive performance. In feedforward networks, a stack of nonlinear transformations, or ${ \{f_\ell(\mathbf{z}_{\ell-1};\theta_\ell)\}^L_{\ell=1} }$, is applied to the input $\mathbf{x}$. As an example, if the network is a multilayer perceptron (MLP), each transformation $\mathbf{z}_\ell = f_\ell(\mathbf{z}_{\ell-1})$ produces an output $\mathbf{z}_\ell$ from the value $\mathbf{z}_{\ell-1}$ of the previous layer with the help of a weight matrix $ \theta_\ell = \{W_{(\ell-1) \rightarrow \ell}\}$. $f_\ell$  is decomposed into two operations (biases omitted for clarity):
\begin{align}
\mathbf{z}_\ell &= \phi_\ell(\mathbf{h}_\ell),\quad \mathbf{h}_\ell = W_{(\ell-1) \rightarrow \ell} \cdot \mathbf{z}_{\ell-1} \label{eqn:mlp}
\end{align}
where $\phi_\ell$ is an activation function,  ${ \mathbf{z}_\ell \in \mathcal{R}^{H} }$ is the post-activation of layer $\ell$ while ${ \mathbf{h}_\ell \in \mathcal{R}^{H} }$ is the pre-activation vector of layer $\ell$.
Note that a matrix multiplication is denoted by $(\circ \cdot \circ)$, a Hadamard multiplication is denoted by $(\circ \otimes \circ)$, and $(\circ)^T$ denotes the transpose operator. For convenience, we set ${ \mathbf{z}_0 = \mathbf{x} }$ (referring to the input vector) and $\mathbf{z}_L$ is the final output or prediction made by the stacked model $f_\Theta(\mathbf{x})$.
We have also introduced special notation for our synaptic weight matrices, where ${ W_{i \rightarrow j} }$ indicates that this parameter matrix connects neurons in layer $i$ to $j$. 

For classification, the output activation is the softmax: $\mathbf{y} ={ \phi_L(\mathbf{v}) = \exp(\mathbf{v}) / ( \sum_j \exp(\mathbf{v}[j]) ) }$, where $j$ indexes scalar elements of a vector.
Any element  in the output vector, i.e., $\mathbf{y}[j]\equiv { \phi_L(\mathbf{v})[j] = p(j|\mathbf{v}) }$, is the scalar probability of class $j$.
Generally, the goal of training is to adjust $\Theta$ to minimize the output loss known as the negative Categorical log likelihood, or ${ L(\mathbf{y}, \mathbf{v}) = -\sum_i \big( \mathbf{y} \otimes \log p(\mathbf{y}|\mathbf{v}) \big)[i] }$.

\subsection{The Learning Algorithm}
\label{sec:algo}
The central idea behind our algorithm, recursive local representation alignment (rec-LRA), is that every layer, not just the output layer, has a target and each layer's parameters/weights are adjusted so that its output moves closer to its target. 
While this idea is also an aspect of prior work such as target-prop \cite{Miguel14,Bengio14,lee2015targetprop}, one key difference between rec-LRA and these prior efforts is that rec-LRA chooses targets that are in the ``possible representation'' of the associated layers. Hence, a layer's parameters are updated more effectively, i.e., a layer is {\em not} forced to match a target that is impossible to achieve
.\footnote{For example, target-prop works by using noise injection (and an encoding-decoding cycle) which could generate targets that ``jump'' far from a layer's current activity (if not carefully controlled by the user), yielding a jolting, high-magnitude weight perturbation that easily results in unstable optimization \cite{ororbia2018biologically}.} 
Furthermore, rec-LRA stands in contrast to recently-explored feedback alignment algorithms \cite{lillicrap2016random,akrout2019weighttransport}, i.e., instead of trying to mimic backprop's way of generating teaching signals (without using forward weights), rec-LRA introduces a special processing unit that creates useful perturbation signals locally. %
Thus, rec-LRA can be viewed either as an alternative to such approaches, and potentially as a complementary technique to most neural design choices, such as residual blocks and other layers that might be helpful for problem-specific representations that a deep network would need to acquire.
Our algorithm, which generalizes ideas in \cite{ororbia2018biologically}
and is grounded in neuro-mechanistic predictive coding theory \cite{rao1999predictive}, aims to break the credit assignment problem into smaller, easier sub-problems that are solvable in parallel of each other.
rec-LRA's goal is to aggressively decompose the underlying directed, acyclic computation graph that defines any stacked neural architecture into small, operation ``sub-graphs''. In this paper, we will show that rec-LRA, through its error synapses, breaks down a network into its $L$ individual transformations, $\{f_\ell(\mathbf{z}_{\ell-1};\theta_\ell)\}^L_{\ell=1}$.
It follows that this divide-and-conquer behavior naturally facilitates distributed training if high performance computing resources are available.

To specify rec-LRA, we start by defining the function it is ultimately meant to optimize, the \emph{total discrepancy}
, which is a ``pseudo-energy function'' that measures the amount of overall system disorder. Specifically, this function computes the degree of mismatch between the current activity of a neural architecture's layers and the activity of a set target activities/states. rec-LRA automatically determines the targets but, in principle, the target could come from external sources or be internally generated based on some partially observed external data, representing values that the network's neuronal processing elements should have taken in order to better predict aspects of its environment.  
Under the framework of discrepancy reduction, a neural system is to minimize the weighted sum of local representational mismatch functions:
\begin{align}
    \mathcal{D}(\Theta) &= \sum^L_{\ell=1} \kappa_\ell  \mathcal{L}_\ell(\mathbf{y}_\ell, \mathbf{z}_\ell), \label{eqn:total_discrepancy} \\ 
    &\mbox{where, } \mathcal{L}_\ell(\mathbf{y}_\ell, \mathbf{z}_\ell) = \big( ||\mathbf{z}_\ell - \mathbf{y}_\ell||_p \big)^q \nonumber
\end{align}
where $\{\mathbf{y}_1,\cdots,\mathbf{y}_\ell,\cdots,\mathbf{y}_L  \}$ are the layer-wise targets and $\mathbf{y}_L$ is the output (i.e. it is $\mathbf{y}$). The value
$p$ sets the type of distance function or norm used to compute mismatch between a state's prediction and the actual target, i.e., $p=2$ is the L2 (Euclidean) norm and $p=1$ is the L1 (Manhattan) norm (typically $q = p$). 
For this study, we set $p = q = 2$ and choose the Euclidean distance function as our representational mismatch function.
The scalar $\kappa_\ell$ is a local coefficient that, while typically set to one for all layers, i.e., $\kappa_1 = \cdots = \kappa_\ell = \cdots = \kappa_L = 1$, if set to values less than one, one could simulate different time-scales of parameter evolution within various levels of the model.  

By taking derivatives of objective in Equation \ref{eqn:total_discrepancy} with respect to each layer of neurons, one can derive vectors of special neurons called ``error neurons'', or $\mathbf{e}_\ell$ (derivation provided in the appendix). 
These neurons measure the difference between the post-activity values of one set of neurons $\mathbf{z}_\ell$ with a corresponding set of target activity values $\mathbf{y}_\ell$. These error units can then be used to form the backbone of a two-phase learning process, using only forward operations: 1) a target generation phase aided by the use of synaptic parameters that transmit mismatch signals across the system, and 2) a local weight update that does not require knowledge of the point-wise derivatives of the ANN's layer-wise activities. 


One particularly powerful and previously unexplored aspect of the discrepancy framework is that the target generation process is not constrained to be symmetrical to the feed-forward phase that computes $f_\Theta(\mathbf{x})$. 
This means that, when conducting credit assignment, error information is not constrained to trace backwards the same pathway taken by the signals that propagated forward through the network during inference. This sharply contrasts with backprop, which requires derivative information to move back along a global feedback pathway that starts from the network's output units back along the same weights used to forward propagate information
, a specific error circuitry that follows from applying the chain rule of calculus to the output cost function. 
This global feedback pathway is not only neuro-biologically implausible but it is the central cause of the well-known vanishing/exploding gradient problem \cite{glorot2010understanding} since a single error signal traversing back  along the central information propagation pathway of $f_\Theta(\mathbf{x})$ is constantly multiplied by the local derivatives of each layer that it passes through. 
In our learning framework, error signals are instead transmitted to the regions/layers of the subgraphs that require them through the use of what we call \emph{skip-error connections}.
Skip-error synapses facilitate a direct transmission of mismatch signals computed by neurons at any layer $i$ directly to any layer $j$, serving as a short-circuit pathway. 
One could also interpret these short-circuit pathways as ``error highways'', inspired by the forward synaptic skip connections used to improve the stability of learning deep ANNs via backprop \cite{srivastava2015highway}.

\begin{figure*}[t]
  \centering
  \begin{tabular}{ >{\centering\arraybackslash}m{6.75cm} >{\centering\arraybackslash}m{6.75cm} }
    \includegraphics[width=0.75\linewidth]{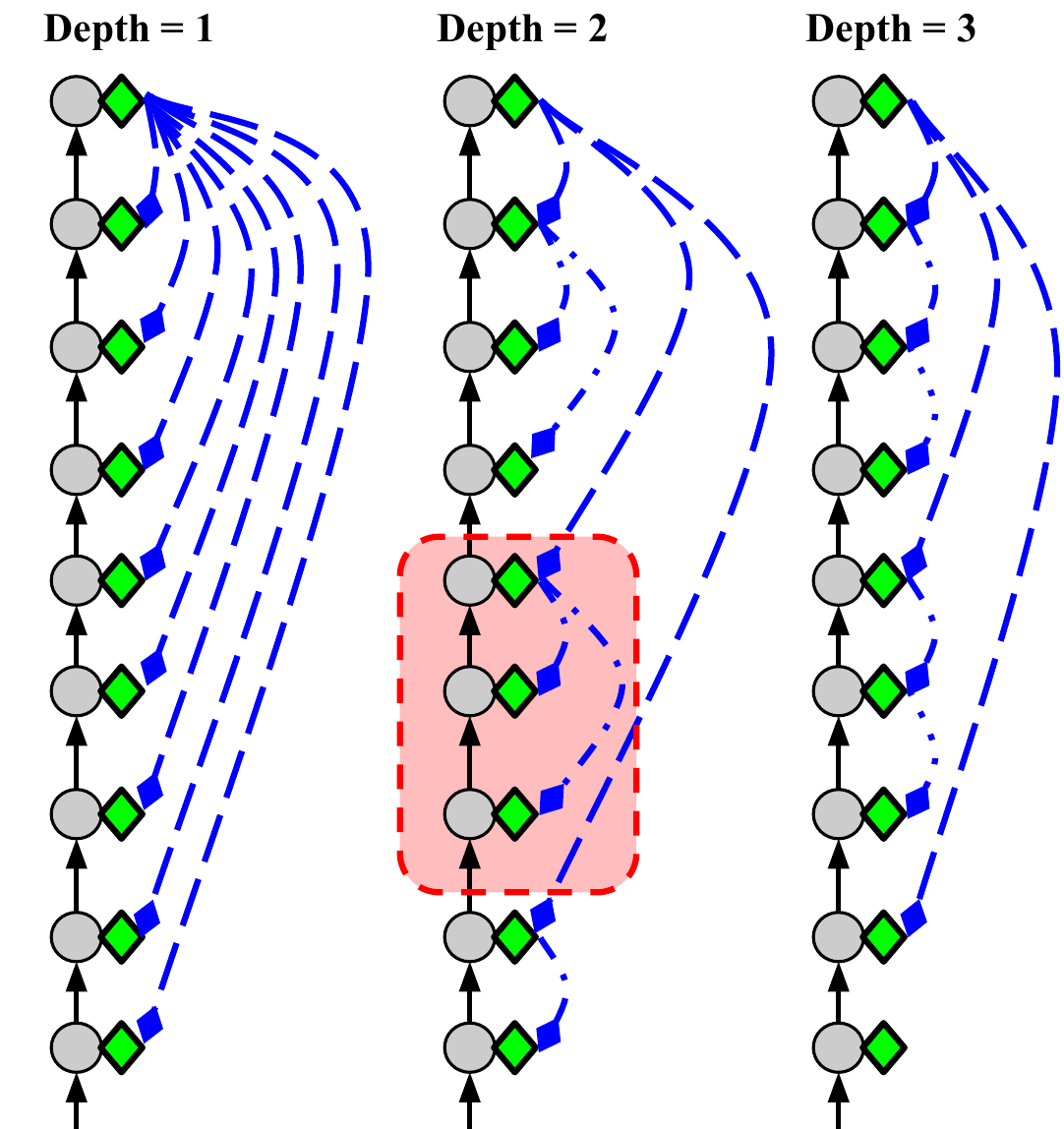} & \includegraphics[width=0.75\linewidth]{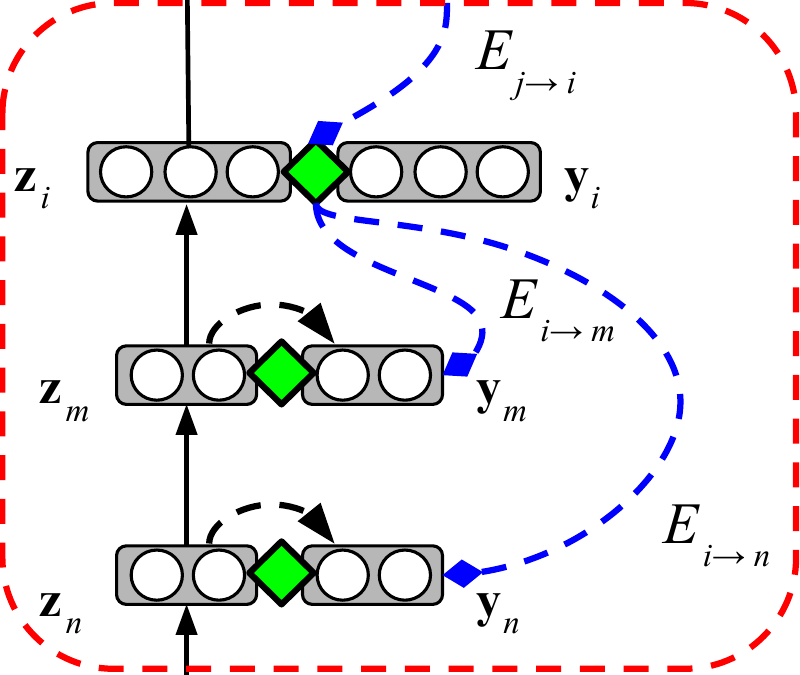} \\ 
    \small (a) Various possible rec-LRA error transmission pathways (ordered left-to-right by recursive depth). & \small (b) Zoom-in of the target generation neural circuit for the area of the architecture highlighted in red box.  \\    
   \end{tabular}
\caption{Recursive local representation alignment error transmission wiring and circuitry.}
\label{fig:wiring}
\vspace{-0.5cm}
\end{figure*}

Under rec-LRA (some error pathways are depicted in Figure \ref{fig:wiring}), targets can be likened to latent representations that would be more desirable when predicting $\mathbf{y}$ to $\mathbf{x}$.  
For a layer $i$ of neurons, a target is computed by taking the mismatch signal computed by error neurons $\mathbf{e}_j$ at layer $j$ in the network and transmitted across a set of error synapses $E_{j \rightarrow i}$, yielding a (vector) displacement signal that communicates to layer $i$'s error units $\mathbf{e}_i$ just how much the layer's activity needs to be adjusted to better please the mapping from $\mathbf{x}$ to $\mathbf{y}$. Formally, the target for layer $i$ is computed as follows:
\begin{align}
    \mathbf{y}_i &= \phi_i( \mathbf{h}_i - \beta \mathbf{d}_i ) &\mbox{// Note: } \mathbf{d}_i = E_{j \rightarrow i} \cdot \mathbf{e}_j \label{eqn:target} \\
    \mathbf{e}_j &= \mathbf{z}_j - \mathbf{y}_j \quad &\mbox{// Assuming $p = q = 2$} \label{eqn:error}
\end{align}
noting that all that is required for computing a target at $i$ is its original pre-activation vector and knowledge of its post-synaptic activation function $\phi_i(\circ)$. $\beta$ is the modulation factor to control the influence of the transmitted displacement message from node $j$ to $i$. Again, notice that we explicitly indicate the direction of transmission from region $j$ to $i$ with the subscript notation $j \rightarrow i$ for error synapses $E_{j \rightarrow i}$.
In an MLP, $\mathbf{h}^i$ would be the pre-activation of a layer $i$ (as in Equation \ref{eqn:mlp}) with the post-activity of that layer $\mathbf{z}^i$ computed by applying a non-linear activation function, such as the linear rectifier, $\phi^i(v) = max(0, v)$, or a non-differentiable function such as the signum, $\phi^i(v) = sign(v)$.
However, $\mathbf{h}^i$ could be the output of a complex function, such as a stack of operations, i.e., convolution and max-pooling operators, as in the case of a residual convolutional network. The error neurons $\mathbf{e}_i$ at layer $i$ then compare the target $\mathbf{y}_i$ to the original activity $\mathbf{z}_i$. 

Once a target for any layer has computed, such as the one for layer $i$ described above, the update for a synapse follows a local Hebbian-like form (the rule can also be derived from the objective, as shown in the appendix). For example, if layer $i$ was connected to an earlier layer/processing stage $k$ by a dense weight matrix $W_{k \rightarrow i}$, then the update using the target representation computed above would be:
\begin{align}
    \Delta W_{k \rightarrow i} = (\mathbf{z}_i - \mathbf{y}_i) \cdot (\mathbf{z}_k)^T = \mathbf{e}_i \cdot (\mathbf{z}_k)^T \label{eqn:fweight_update}
\end{align}
where $(\circ)^T$ denotes the transpose operation. If the layers $i$ and $k$ were related by something other than a dense matrix, such as a set of filters or noise maps, the update rule could be readily adjusted to deal with the operation under question (for example, the rule would be follow the form provided in the section on pseudo-convolution).
The error synaptic weight matrix $E_{j \rightarrow i}$ that relayed information from layer $j$ to $i$ is also updated using a local Hebbian-like rule:
\begin{align}
    \Delta E_{j \rightarrow i} = \gamma(-\mathbf{d}_i \cdot (\mathbf{e}_j)^T) \label{eqn:eweight_update}
\end{align}
where $\gamma$ is scalar factor for controlling the strength of the error synaptic adjustment (a value less than $1$ means the error weights change more slowly than the forward weights).

So armed with the perspective above, rec-LRA as a general procedure would first run the forward pass procedure of $f_\Theta(\mathbf{x})$ and then compute the targets and mismatch signals (Equations \ref{eqn:target} \& \ref{eqn:error} ) which can immediately be used for weight update calculations (Equations \ref{eqn:fweight_update} \& \ref{eqn:eweight_update}). The inherent parallelism in the target and mismatch computations stems from the fact that the error pathway need not be symmetrical to the forward transmission pathway. If a point $j$ had error synapses connecting to points/neighbors $m$ and $n$, the error transmission to each point could happen in parallel since the displacement calculation at $n$ does not depend on that conducted at $m$. This means that transmission of mismatch signals to each of $j$'s neighbors can be done on separate processors, if available. 
In Figure \ref{fig:wiring}, this type of error transmission circuitry is graphically depicted ( Figure \ref{fig:wiring}b).

For a feedforward architecture, rec-LRA would start operating at the output layer $L$, then compute the targets for the inner regions that the error neurons at $L$ connect to, and then recursively call itself on each of those target regions, subsequently computing the appropriate error neuron vectors and further computing targets for regions that connect to those regions, and so on and so forth. The base case for the recursion's termination would simply be when it encounters regions that do not immediately connect to anywhere else. This is formally depicted in rec-LRA's architecture-agnostic algorithmic form, the full details of which are provided in the appendix. It is important to note that the weight matrices (both $W_{k \rightarrow i}$ \& $E_{j \rightarrow i}$) that connect to a region $i$ can be readily updated as soon as the local error neuron signals are available. To truly exploit the potential speed offered by rec-LRA's parallel nature, one could allocate each recursive call to a cluster/set of CPUs/GPUs dedicated to generating targets/updates for various parts of the operator graph.

While rec-LRA works with any neural architecture, one may note that the design of its recurrent error pathways is left up to the modeler, who might care to investigate how model generalization changes with respect to error wiring.
To ease the design process, a modeler could automate this design choice by employing an outer search method, e.g., neural architecture search \cite{elsken2018neural}, or 
could craft simple pathways by taking into account the dimensionality of the network's various layers and the number of processors available.\footnote{We analyze error wiring pattern choices in the appendix.} For example, one might choose the middle wiring pattern (the depth 2 model) in Figure \ref{fig:wiring} if layers $2$, $5$, and $8$ are bottleneck layers (which contain a low number of neurons) and $5$ GPUs are available -- rec-LRA would first use $3$ GPUs to parallel compute targets/mismatches for layers $8$, $5$, and $2$ and then $5$ GPUs to parallel compute the targets/mismatches for the remaining layers ($1,3,4,6,7$). Alternatively, if one is using a special architecture with repeating design ``blocks'' such as transformer for language processing \cite{devlin2018bert} or residual network \cite{he2016deep} for image processing (studied in the next section), one could use the model's natural grouping of processing layers to create a natural error transmission pathway. 

\subsection{Residual Neural Networks and rec-LRA}
\label{sec:resnets}
Residual neural networks (ResNets) \cite{he2016deep,he2016identity}, recently reaching state-of-the-art performance on popular vision benchmarks, are architectures that are composed of many hidden layers wired together with a special forward connectivity pattern. 
Specifically, residual networks utilize skip/shortcut connections that allow the forward propagation of information to jump over some hidden layers, specifically those that might not prove useful in mapping $\mathbf{x}$ to $\mathbf{y}$. 
Formally, the layers in the network that permit a residual mapping are defined as: $\mathbf{z}_\ell = f_\ell(\mathbf{z}_{\ell-1}; \theta_\ell) + \mathbf{z}_{\ell-g}$
where $g$ controls the length of the gap/skip, typically of size $2$ or $3$. The idea behind the formulation above is that, in the event that directly fitting the transformation function $f_\ell(\mathbf{z}_{\ell-1}; \theta_\ell)$ is too challenging, the residual mapping (as indicated by the second term of the equation) will be easier to optimize. 
This gives the network the choice of retaining the input if it finds that a particular layer(s) is not needed. The transformation $f_\ell(\mathbf{z}_{\ell-1};\theta_\ell)$ could range from being a linear transformation to a stack of fully-connected layers (as in Equation \ref{eqn:mlp}). 
In computer vision, it is often formulated as a residual ``block'', i.e., a stack of operations including convolutions, the relu activation ($\mathbf{v} = max(0, \mathbf{v})$), pooling, normalization layers, etc (the block we used is shown in appendix).

Training a residual network with rec-LRA exploits the block-based structure of the network to craft the error message transmission pathways.
If, for example, a residual block is a stack of nonlinear transformations, we can choose to embed a vector of error neurons at the output of each residual block and wire them to the output error neurons at layer $L$. In the case of the two residual blocks depicted in For the first level of recursion, we would wire the output layer $L$ directly (via $E_{L \rightarrow i}$) to any residual block output vector $\mathbf{z}_i$.
\begin{wrapfigure}{r}{3.2in}
\begin{minipage}{0.5\textwidth}
\vspace{-0.75cm}
\begin{algorithm}[H]
\caption{Rec-LRA (depth 2) applied to L-layered network $f_\Theta(\mathbf{x})$ with residual gap $g$.}
\label{algo:rec_lra_resnet}
\begin{algorithmic}[1]
    \State {\bf Inputs:} $\mathbf{x}$, $\mathbf{y}$, $\Theta = \{W_1,\cdots,W_L \}$, $\beta$, $\gamma$, $g$
    \State \hspace{0.1cm}  $\Theta_E = \{ E_{L \rightarrow (L-1)}, \cdots, E_{i \rightarrow j}, \cdots, E_{L \rightarrow 1} \}$
    \LineComment{Inference procedure for network $f_\Theta(\mathbf{x})$}
    \Function{RunModel}{$\mathbf{x}$, $\Theta$} 
         \For{$\ell = 1$ to $L$}
            \If{$\ell \mod g \equiv 0$}
                \State $\mathbf{h}_\ell = W_\ell \cdot \mathbf{h}_{\ell-1} + \mathbf{z}_{\ell-g}$
            \Else
                \State $\mathbf{h}_\ell = W_\ell \cdot \mathbf{h}_{\ell-1}$
            \EndIf
            \State $\mathbf{z}_\ell = \phi_\ell(\mathbf{h}_\ell)$
         \EndFor
         \State \Return $\mathcal{Z} = \{\mathbf{z}_0, \mathbf{z}_1, \cdots, \mathbf{z}_L \}$
    \EndFunction
    \LineComment{Compute error neurons given activities}
    \Function{CalcErrNeurons}{$\mathbf{y}$, $\Theta$, $\Theta_E$, $\mathcal{Z}$}
        \State $\mathbf{y}_L = \mathbf{y}$,\quad $\mathbf{e}_L = \mathbf{z}_L - \mathbf{y}_L$
        \For{$\ell = (L-1)$ to $1$}
            \If{$\ell \mod g \equiv 0$}
                \State $\mathbf{d}_\ell = E_{L \rightarrow \ell} \cdot \mathbf{e}_{L}$
            \Else
                \State $\mathbf{d}_\ell = E_{\ell+1 \rightarrow \ell} \cdot \mathbf{e}_{\ell+1}$
             \EndIf
             \State $\mathbf{y}_\ell = \phi_\ell(\mathbf{h}_\ell - \beta \mathbf{d}_\ell)$,\quad $\mathbf{e}_\ell = \mathbf{z}_\ell - \mathbf{y}_\ell$
        \EndFor
        \State \Return $\mathcal{E} = \{\mathbf{e}_1, \cdots, \mathbf{e}_L \}$
        \State \hspace{0.9cm} $\Upsilon = \{\mathbf{d}_1, \cdots, \mathbf{d}_L \}$
    \EndFunction
    \LineComment{Compute updates given error neurons} 
    \Function{ComputeUpdates}{$\mathcal{E}$, $\mathcal{Z}$, $\Upsilon$}
        \For{$\ell = 1$ to $L$}
            \State $\Delta W_\ell = \mathbf{e}_\ell \cdot (\mathbf{z}_{\ell-1})^T$
            \If{$\ell > 1$}
                \If{$\ell \mod g \equiv 0$}
                    \State $\Delta E_{L \rightarrow \ell} = -\gamma(\mathbf{d}_{\ell-1} \cdot (\mathbf{e}_{L})^T$)
                \Else
                    \State $\Delta E_{\ell+1 \rightarrow \ell} = -\gamma(\mathbf{d}_{\ell-1} \cdot (\mathbf{e}_{\ell})^T$)
                \EndIf
            \EndIf
        \EndFor
        \State \Return $\{\Delta W_1, \Delta W_2, \Delta E_2, \cdots, \Delta W_L, \Delta E_L \}$
    \EndFunction
\end{algorithmic}
\end{algorithm}
\end{minipage}\vspace{-0.5cm}
\end{wrapfigure}
Wiring skip-error connections in this way means that rec-LRA treats each residual block as a computational subgraph (which maps a representation $\mathbf{z}_{i-g}$ to $\mathbf{z}_{i}$). Once a skip-error connections wired to each a block generates its desired target, rec-LRA will recursively enter the block to compute its internal error neurons and weight updates, independently of the blocks above and below, effectively decoupling its update calculation from the rest of the blocks.
In treating the residual blocks as decoupled computation graphs, one could view the output of each block as a ``meta-representation'' (in Figure \ref{fig:wiring}a, for the middle model, these would be layers $2$, $5$, and $9$) , or a post-activation layer that serves as the first focus of LRA's target generation process. The other layers within it
serve as computational ``support'' layers (layers $1,3,4,6,7$ in Figure \ref{fig:wiring}a).
Algorithm \ref{algo:rec_lra_resnet} formally depicts how rec-LRA operates on a (fully-connected) residual network with skip $g$.

While rec-LRA in the form we have proposed works strictly with strictly neuro-cognitively plausible learning rules (Equations \ref{eqn:fweight_update}, \ref{eqn:eweight_update} as used in Algorithm \ref{algo:rec_lra_resnet}), one could opt to mix other learning algorithms with the rec-LRA framework. For instance, one could use rec-LRA to generate meta-representation targets for any residual block output $\mathbf{z}_i$ and employ a procedure like backprop (treating the block's output error neuron vector as a proxy for $\frac{\partial \mathcal{L}}{\partial \mathbf{z}_i}$) or Hebbian rules \cite{hebb1949organization} to compute local weight updates (reducing the number of error matrices needed and saving on memory).

\noindent \textbf{Replacing Convolution with Fixed Perturbation:}
To further save on computation, we replaced the convolutional operator with a fixed noise ``pseudo-convolution'', which was proposed in \cite{juefei2018perturbative} (referred to as a ``perturbative layer''). As was shown in \cite{juefei2018perturbative}, the pseudo-convolution is not only drastically faster than actual convolutional but the generalization performance of the underlying model using it is comparable to one with convolution. A pseudo-convolution is computed as follows:
\begin{align*}
    \mathbf{z}^c_\ell &= \sum^M_{m=1} w^m_\ell \phi_r( \mathbf{z}^m_{\ell-1} + \mathbf{n}^m_{\ell-1}), \mbox{where, } \phi_r(v) = \max(0,v)
\end{align*}
where $w^\ell_m$ is a scalar weight that is applied to its corresponding noise map.
In the above formula, we see that $M$ noise maps $\mathbf{n}^m_\ell$ must be cycled through in order to compute the final desired output channel (map) $\mathbf{z}^\ell_c$. The idea is that, for the price of the memory required to store the pre-generated $M$ noise maps (the elements of each are each sampled from a centered Gaussian distribution, $\sim \mathcal{N}(0,\sigma^2)$), we side-step the need for learn-able kernel parameters for the convolution operation (cutting out another convolution per filter update).
The only parameters in a pseudo-convolution that require updating are the linear combination weights ($1$ update per scalar weight applied to each noise map).
Under rec-LRA, which would embed error neurons right next to $\mathbf{z}^c_\ell$, the update for the $m$th noise map weight $w^\ell_m$ would be:
\begin{align*}
    \Delta w^m_\ell = \sum_i \sum_j \Big( \mathbf{e}_\ell * \big( \phi_r( \mathbf{z}^m_{\ell-1} + \mathbf{n}^m_{\ell-1}) \big)^T \Big) [i,j]
\end{align*}
where the update is collapsed by summing over all dimensions to get a scalar update for weight $w^m_\ell$.

\begin{figure*}[t]
\centering
  \includegraphics[width=0.3275\linewidth]{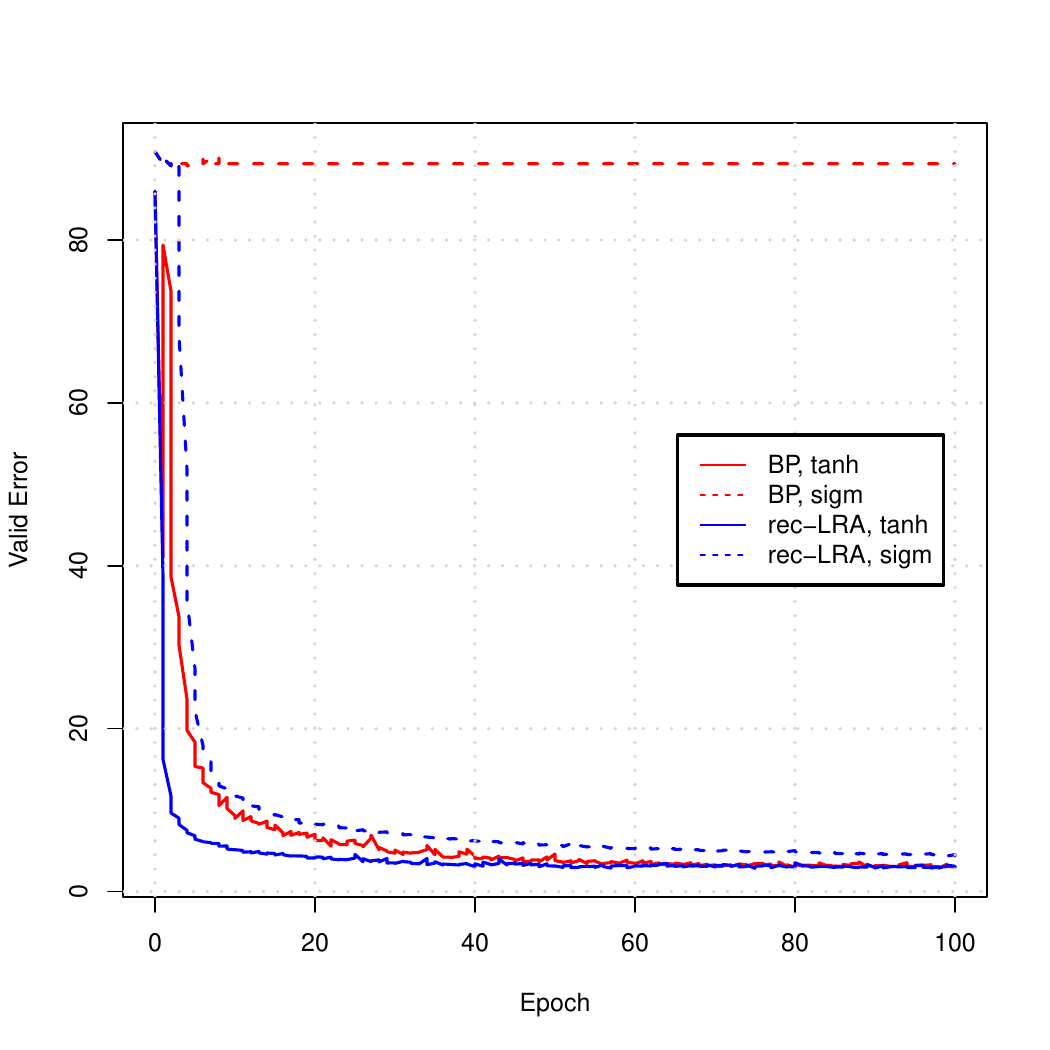}
  \includegraphics[width=0.3275\linewidth]{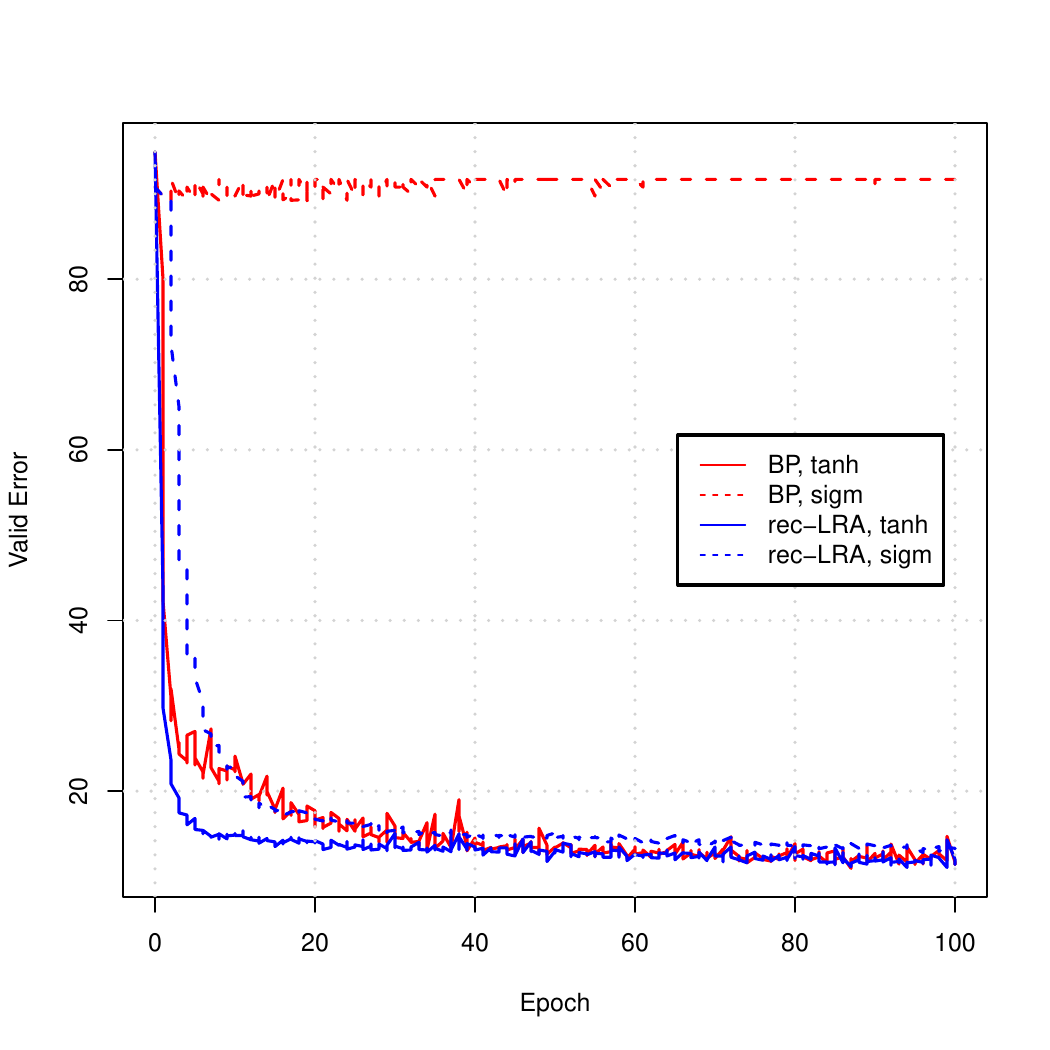}
  \includegraphics[width=0.3275\linewidth]{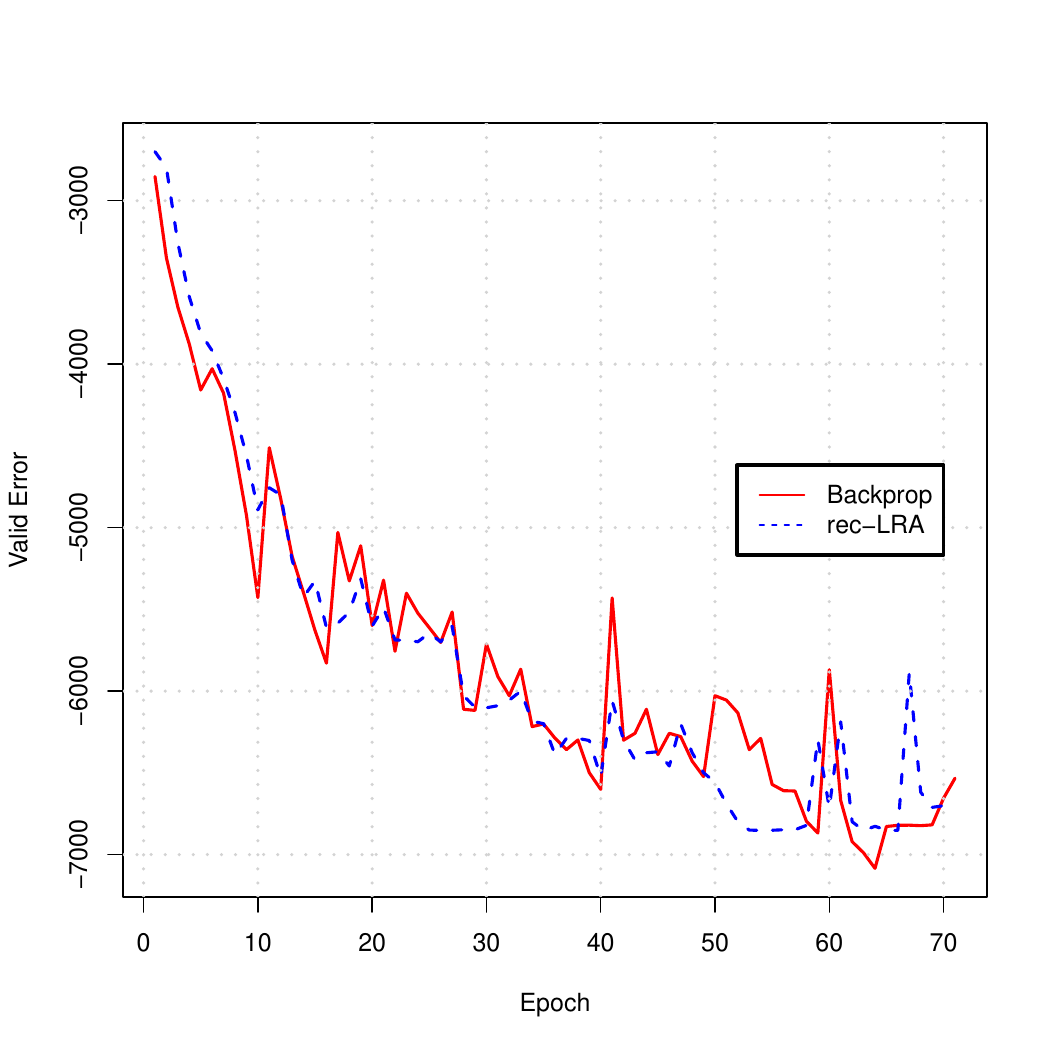}
  \caption{Error for networks on MNIST (top), FMNIST (middle), and ImageNet (with ResNets) (right).}
  \label{fig:convergence}
\end{figure*}

\section{Experiments}
\label{sec:exp}
We experiment with the proposed rec-LRA and compare it to results reported for other backprop-alternatives. Specifically, we adapted rec-LRA to fully-connected MLPs, convolutional networks (CNNs), and residual networks (ResNet). Further experimental details can be found in the appendix.

\noindent \textbf{MNIST \& Fashion MNIST:} This dataset 
contains $28\times28$ images with gray-scale pixel feature values in the range of $[0,255]$. The only preprocessing applied to this data is to normalize the pixel values to the range of $[0,1]$ by dividing them by 255.
On the other hand, Fashion MNIST (FMNIST) \cite{xiao2017fashion} serves as a challenging drop-in replacement for MNIST. Fashion MNIST (pre-processed the same as MNIST) contains images each depicting one of $10$ clothing items. 
Training had $60000$ samples, testing had $10000$, and $2000$ validation samples was drawn from the training set.
In Table \ref{results:benchmarks}, we report our classification error on both training and test sets for rec-LRA and compare to prior reported results. Prior results have been reported for backprop (BP) as well as relevant biologically-motivated, gradient-free algorithms such as feedback alignment (FA), direct feedback alignment (DFA), error-driven local representation alignment (LRA-E), equilibrium propagation (E-Prop), and target propagation (TP) (\cite{bartunov2018assessing} \& \cite{ororbia2018biologically}). 
For the rec-LRA results, we report 4 variations ($5$ layers, $256$ units), each using a different activation function. To be comparable to prior work, the first variant of rec-LRA utilizes hyperbolic tangent units (rLRA, tanh). 
The next two variants used were linear rectifier units (rLRA, relu) and exponential linear units (rLRA, elu) to demonstrate compatibility with various popular activation functions. Finally, signum units (rLRA, sign) were tested in order to investigate rec-LRA's ability to train networks with non-differentiable functions.
We observe that rec-LRA outperforms all of the other algorithms on FMNIST, including backprop. On MNIST, rec-LRA outperforms all of the other gradient-free alternatives but does not beat out backprop. While the signum networks do not reach the performance of the topmost networks, they are not among the worst performing, which offers encouraging evidence that non-differentiable networks can make viable classifiers.
We further analyzed the training dynamics of more complex, nonlinear networks, i.e., $8$ layers of either $256$ logistic sigmoid or tanh neurons, trained via backprop and rec-LRA over $100$ epochs. Deep sigmoidal models are known to be very difficult to train due to the well-known vanishing gradient problem \cite{glorot2010understanding}, especially if naive Gaussian initialization is used.
In Figure \ref{fig:convergence}, on both MNIST and FMNIST, we observe that rec-LRA successfully trains networks of both kinds of units with the same initialization and converges sooner. The fact that this result holds for the networks with tanh units, which are friendlier to a backprop-centric optimization, offers some evidence of rec-LRA's potential robustness and stability.

\begin{table}[t]
\begin{center}
\parbox{.5\linewidth}{
\centering
\caption{Classification error on MNIST \& FMNIST.}
\label{results:benchmarks}
 \begin{tabular}{l | c c c c} 
 \hline
    & \multicolumn{2}{c}{MNIST}  & \multicolumn{2}{c}{FMNIST} \\ 
 \textbf{Algorithm} & Train & Test & Train & Test\\ 
 \hline
 BP & $\bf 0.00$ & $\bf 1.48$ & $\bf 12.10$ & $\bf 12.98$ \\
 TP & $\bf 0.00$ & $\bf 1.86$ & $21.078$ & $19.66$ \\
 E-Prop  & $7.59$ & $9.21$ & $16.56$ & $20.97$ \\
 LRA-E  & $0.16$ & $1.97$ & $\bf 9.84$ & $\bf 12.31$ \\
 FA & $\bf 0.00$ & $\bf 1.85$ & $\bf 12.09$ & $\bf 12.89$ \\
 DFA & $0.85$ & $2.75$ & $12.58$ & $13.09$ \\
 \hline
 rLRA, tanh & $\bf 0.00$ & $\bf 1.82$ & $\bf 6.57$ & $\bf 11.87$ \\
 rLRA, relu & $0.22$ & $2.26$ & $8.95$ & $14.13$ \\
 rLRA, elu & $0.09$ & $1.93$ & $9.39$ & $13.17$ \\
 rLRA, sign & $0.85$ & $2.33$ & $12.42$ & $14.88$ \\
 \hline
\end{tabular} 
}
\hfill
\parbox{.45\linewidth}{
\centering
 \caption{CIFAR-10 generalization error.}
\label{results:cifar10}
 \begin{tabular}{l | c c } 
 \hline
 \textbf{CIFAR-10} & Train & Test \\ 
 \hline
 TP & $28.69$ & $39.47$ \\
 FA  & $17.46$ & $37.44$ \\
 DFA & $32.74$ & $44.41$ \\
 \hline
 CNN-BP  & $7.89$ & $33.17$ \\ 
 CNN-rLRA & $13.88$ & $35.22$ \\ 
 ResNet-BP  & $\bf 5.00$ & $\bf 5.94$ \\ 
 ResNet-rLRA & $\bf 5.88$ & $\bf 6.12$ \\ 
 \hline
 \end{tabular}
}
\vspace{0.15cm}
 \caption{CIFAR-10 convolution (conv) vs. psuedo-convolution (pconv) validation error. }
 \label{results:cifar_conv_v_pnn}
  \begin{tabular}{l | c c } 
  \hline
  \textbf{CIFAR-10} & Valid Err & Best Epoch \\
  \hline
  cResNet-BP  & $6.22$ & 125 \\
  pResNet-BP  & $5.95$ & 105  \\
  cResNet-rLRA & $8.80$ & 120   \\
  pResNet-rLRA & $6.12$ & 104 \\
 \hline
 \end{tabular}
\end{center}
\vspace{-0.8cm}
\end{table}

\noindent \textbf{CIFAR-10:} The CIFAR-10 dataset has $50,000$ training and $10,000$ test images, across $10$ categories.  Images are of size $32\times32$ pixels. $5,000$ training samples were set aside to measure validation metrics. Global contrast normalization and ZCA whitening were used to pre-process images.
While this dataset is far more challenging than that of MNIST, we observe in Table \ref{results:benchmarks} that rec-LRA outperforms networks trained with other gradient-free methods, i.e, target prop and feedback alignment. Furthermore, rec-LRA comes quite close to the performance of same model trained with backprop, offering evidence of its ability to handle a challenging color image problem. In the appendix, we further dissect the networks' predictions and visualize latent representations.

To test how performance would compare when using either convolution (conv) or pseudo-convolution (pconv), further classification experiments were conducted on CIFAR-10. We use  ResNet-18  with $64$ perturbation masks for using pconv (pResNet) and  $64$ filters for vanilla Resnet (cResNet). All models used a skip $g = 2$. 
The initial learning rate was set to $10^{-3}$ and reduced by a factor of $10$ at epoch $60$, and later again at epoch $90$ , $120$ and $150$.
We trained models with varying batch sizes and report validation error (Table \ref{results:cifar_conv_v_pnn}), recording the number of epochs required for each model to achieve optimal results. As seen in Table \ref{results:cifar_conv_v_pnn}, we found that the performance difference in using pconv over conv was negligible across batch sizes (but gained a small speed-up of roughly $0.73$ seconds per mini-batch -- note that tuned implementations of $1x1$ conv would change the speed-up).

\noindent \textbf{ImageNet:} The large-scale benchmark ImageNet \cite{russakovsky2015imagenet}, specifically the ILSVRC-2010 subset, contains over $1.2$ million images, of size $224\times224$, each contain one out of $1000$ different categories. Given that the number
\begin{wraptable}{r}{2.38in}
\vspace{-0.365cm}
\caption{ImageNet generalization error.}
 \label{results:imagenet}
 \begin{tabular}{l | c c} 
 \hline
 \textbf{ImageNet} & Top-1 & Top-5 \\ 
 \hline
 CNN, TP & $98.34$ & $94.56$ \\
 CNN, FA  & $93.08$ & $82.54$ \\
 ResNet, FA+BP  & $73.01$ & $51.24$ \\
 ResNet, SS \cite{xiao2018biologically} & $37.91$ & $16.18$ \\
 ResNet, SS+BP \cite{xiao2018biologically} & $37.01$ & $15.44$ \\ ResNet, FA+WM \cite{akrout2019weighttransport} & $\mathbf{30.20}$ & N/A \\
 ResNet, KP \cite{akrout2019weighttransport} & $\mathbf{29.2}$ & N/A \\
 \hline
 CNN, BP & $62.58$ & $39.89$ \\ 
 CNN, rLRA & $73.69$ & $49.78$ \\ 
 ResNet, BP & $\mathbf{28.15}$ & $\mathbf{9.81}$  \\
 ResNet, rLRA  & $\mathbf{30.48}$ & $\mathbf{11.97}$ \\
 \hline
 \end{tabular}
\vspace{-0.375cm}
\end{wraptable}
of classes is large, it is convention to report two types of error rates: top-1 and top-5. 
The top-5 error rate is the fraction of test images for which the correct label is not among the $5$ classes considered most probable by the evaluated model.
In Table \ref{results:imagenet}, we observe that rec-LRA-trained models outperformed ones trained via other gradient-free methods and comes quite close to the performance of the backprop-trained architecture (both top-1/top-5 test error). 

Furthermore, we measured wall-clock training time for both networks to determine if rec-LRA training offered a speed-up even though we implemented it in simulation without distributed computing hardware (rec-LRA would run dramatically faster with parallelization). Notably, in terms of total training run-time over $90$ epochs using a small set of $8$ V100 GPUs, the backprop ResNet took $3$ hours and $45$ minutes (min) to train (speed was about $2.5$-$2.7$ min/epoch) while rec-LRA took $2.127$ min/epoch, training over the course of $3$ hours and $12$ min. In Figure \ref{fig:convergence} (right), we see that rec-LRA does reach lower validation error sooner than backprop (though this result is not as obvious as it was for MNIST/FMNIST). Furthermore, rec-LRA converges more smoothly than a backprop-trained ResNet.
We also report the performance of the (best-performing) sign symmetry (SS) of \cite{xiao2018biologically}, which we outperform though the margin of improvement is far narrower (note that SS also uses partial backprop), weight mirrors (WM), \cite{akrout2019weighttransport}, and the classical Kolen-Pollack (KP) algorithm. 
State-of-the-art performance of deep networks on ImageNet is better \cite{xie2019self} than that obtained by gradient-free algorithms such as our own and in \cite{bartunov2018assessing}. However, our aim was to show that a gradient-free algorithm can generalize on difficult, large-scale datasets -- modern-day heuristics would further boost our model performance. 
\vspace{-0.3cm}

\section{Conclusions}
\label{sec:conclusions}
In this paper, we proposed a gradient-free learning algorithm, recursive local representation alignment (rec-LRA), for training deep neural architectures. rec-LRA  generalizes as well as backprop and outperforms other current gradient-free procedures across several datasets, notably on the massive-scale benchmark ImageNet. Furthermore, it offers improved convergence due to faster, parallelizable weight updates, as shown in our experiments. As a result, this work offers empirical evidence that a backprop-free procedure can indeed scale up to larger datasets.


\bibliographystyle{acm}
\bibliography{ref}

\newpage
\vspace{-3cm}

\section*{Derivation of Recursive LRA Model Updates}
\label{sec:derivation}
An artificial neural system, under our proposed framework of discrepancy reduction, is engaged with minimizing the weighted sum of local representational mismatch functions. In general, for a neural system to work effectively under this framework, two neural computational processes must be specified: 1) target representation generation (in order to compute layer representation mismatch errors), and 2) synaptic weight updating (in order to improve the model's long-term performance at guessing target representations that facilitate a good mapping between $\mathbf{x}$ and $\mathbf{y}$) \cite{ororbia2018biologically}. 
Under recursive LRA (rec-LRA), the specification of both computational processes centers around the introduction of a new type of neuron called the ``error neuron'', a processing element solely tasked with calculating mismatch values between the latent states of the network (given data) and target states that better describe an effective mapping between input $\mathbf{x}$ and output (target) $\mathbf{y}$.

To design the error neurons needed for rec-LRA to work, we start from the objective function that a neural system attempts to minimize under the framework of discrepancy reduction.
The objective function, called \emph{total system discrepancy} or \emph{total discrepancy}, is formally defined as:
\begin{align*}
    \mathcal{D}(\Theta) &= \sum^L_{\ell=1} \kappa_\ell  \mathcal{L}_\ell(\mathbf{y}_\ell, \mathbf{z}_\ell),  
    \mbox{where, } \mathcal{L}_\ell(\mathbf{y}_\ell, \mathbf{z}_\ell) = \big( ||\mathbf{z}_\ell - \mathbf{y}_\ell||_p \big)^q 
\end{align*}
where $\{\mathbf{y}_1,\cdots,\mathbf{y}_\ell,\cdots,\mathbf{y}_L  \}$ are the layer-wise targets, $\{\mathbf{z}_1,\cdots,\mathbf{z}_\ell,\cdots,\mathbf{z}_L  \}$ are the current latent states (given input data $\mathbf{z}_0 = \mathbf{x}$), $\{\mathbf{h}_1,\cdots,\mathbf{h}_\ell,\cdots,\mathbf{h}_L  \}$ are their corresponding pre-activation values, and $\mathbf{y}_L$ is the output target (i.e. it is $\mathbf{y}$, the encoded classification label). The value
$p$ sets the type of distance function used to compute mismatch between a state's prediction and the actual target, i.e., $p=2$ is the L2 (Euclidean) norm and $p=1$ is the L1 (Manhattan) norm (typically $q = p$). 
In this work, we set $p = q = 2$, as mentioned in the main paper, choosing the square of the Euclidean distance as our representational mismatch function.
$\kappa_\ell$ is a scalar coefficient used to weight a particular local loss's (at $\ell$) contribution to total discrepancy and, if set to values less than one, could be used to simulate different time-scales of parameter evolution within various levels of the neural system. In the event that $p = q = 2$, the local loss could further be interpreted as a local Gaussian log likelihood where the $\kappa_\ell = \frac{1}{\sigma^2}$ is used to set its fixed scalar variance $\sigma^2$. 
The error neurons themselves are then set to be equal to the partial derivatives of the function with respect to its latent states, or $\frac{\partial \mathcal{D}(\Theta)}{\partial \mathbf{z}_\ell}$. The needed partial derivative for any layer $\ell$ (with $p=q=2$) would be:
\begin{align*}
    \mathbf{e}_\ell = \frac{\partial \mathcal{D}(\Theta)}{\partial \mathbf{z}_\ell} &= \frac{\partial \kappa_\ell \big( ||\mathbf{z}_\ell - \mathbf{y}_\ell||_2 \big)^2 }{\partial \mathbf{z}_\ell} \\
    &= \frac{\kappa_\ell}{2} (\mathbf{z}_\ell - \mathbf{y}_\ell) \otimes \frac{\partial (\mathbf{z}_\ell - \mathbf{y}_\ell)}{\partial \mathbf{z}_\ell} \\
    &= (\mathbf{z}_\ell - \mathbf{y}_\ell), \: \mbox{with } \kappa_\ell = 2 \mbox{.}
\end{align*}
It should be noted that the other kinds of error neurons could be designed to specify rec-LRA's central computational processes using other flavors of local representational mismatch losses.

If we assume a simple feedforward process for the neural system's propagation of information from $\mathbf{x}$ to $\mathbf{y}$, i.e., $\mathbf{z}_\ell = \phi_\ell(\mathbf{h}_\ell)$ and $\mathbf{h}_\ell = W_\ell \cdot \mathbf{z}_{\ell-1}$ which means that the forward parameters are $\{W_1,W_2,\cdots,W_L\}$, then deriving the weight update proceeds from the error neuron derivation in the following manner:
\begin{align}
    \Delta W_\ell &= \frac{\partial \mathcal{D}(\Theta)}{\partial W_\ell} = \frac{\partial \mathcal{D}(\Theta)}{\partial \mathbf{z}_\ell} \frac{\partial \mathbf{z}_\ell}{\partial W_\ell} \nonumber\\
    &= \frac{\partial \mathcal{D}(\Theta)}{\partial \mathbf{z}_\ell} \frac{\partial \mathbf{z}_\ell}{\partial \mathbf{h}_\ell} \frac{\partial \mathbf{h}_\ell}{\partial W_\ell} \nonumber\\
    &= \big( \mathbf{e}_\ell \otimes \phi^{\prime}_\ell(\mathbf{h}_\ell) \big) \cdot (\mathbf{z}_{\ell-1})^T \label{eq:to_approx}
\end{align}
where we observe that the weight updates directly follow from the error neuron derivation.

Note that prior work \cite{ororbia2018conducting, ororbia2018continual} has found that it is permissible to omit the activation function's point-wise derivative so long as the activation is monotonically non-decreasing in its input. This has been demonstrated to work well in several prior efforts \cite{ororbia2018biologically, melchior2019hebbian}. One hypothesis for this empirical performance is that if the input is in the approximately linear region of the activation function, removing the derivative makes no difference; meanwile, if the input is in the saturated region (e.g., left side of ReLU or both tails of the the sigmoid), the activation derivative strongly attenuates the directional signal provided by other parts of the chain rule (especially when multiplied by the 0 derivative in parts of the ReLU function). In such a case, removing the derivative activation would allow a neuron to take a larger step size and escape the saturated region if necessary.
Thus we replace Equation \ref{eq:to_approx} with the update $ \mathbf{e}_\ell \cdot (\mathbf{z}_{\ell-1})^T $
which is also a type of error-driven Hebbian learning rule \cite{hinton1988learning,ororbia2018biologically}, similar in spirit to the classical delta \cite{widrow1960adaptive} and prescribed error rules \cite{macneil2011pes,bekolay2013simultaneous}.

The final remaining part is to define how the targets are generated, i.e., $\{\mathbf{y}_1,\cdots,\mathbf{y}_\ell,\cdots,\mathbf{y}_L  \}$, since the error neurons expect to be provided with some type of target representation that they can use when measuring representational mismatch. While the targets could come from a variety of sources, e.g., the outputs of other complementary neural systems, a database of desired latent representations, or an iterative inference process \cite{ororbia2018conducting}, 
\begin{wrapfigure}{r}{3in}
\begin{center}
\vspace{-1.2cm}
\includegraphics[width=0.475\textwidth]{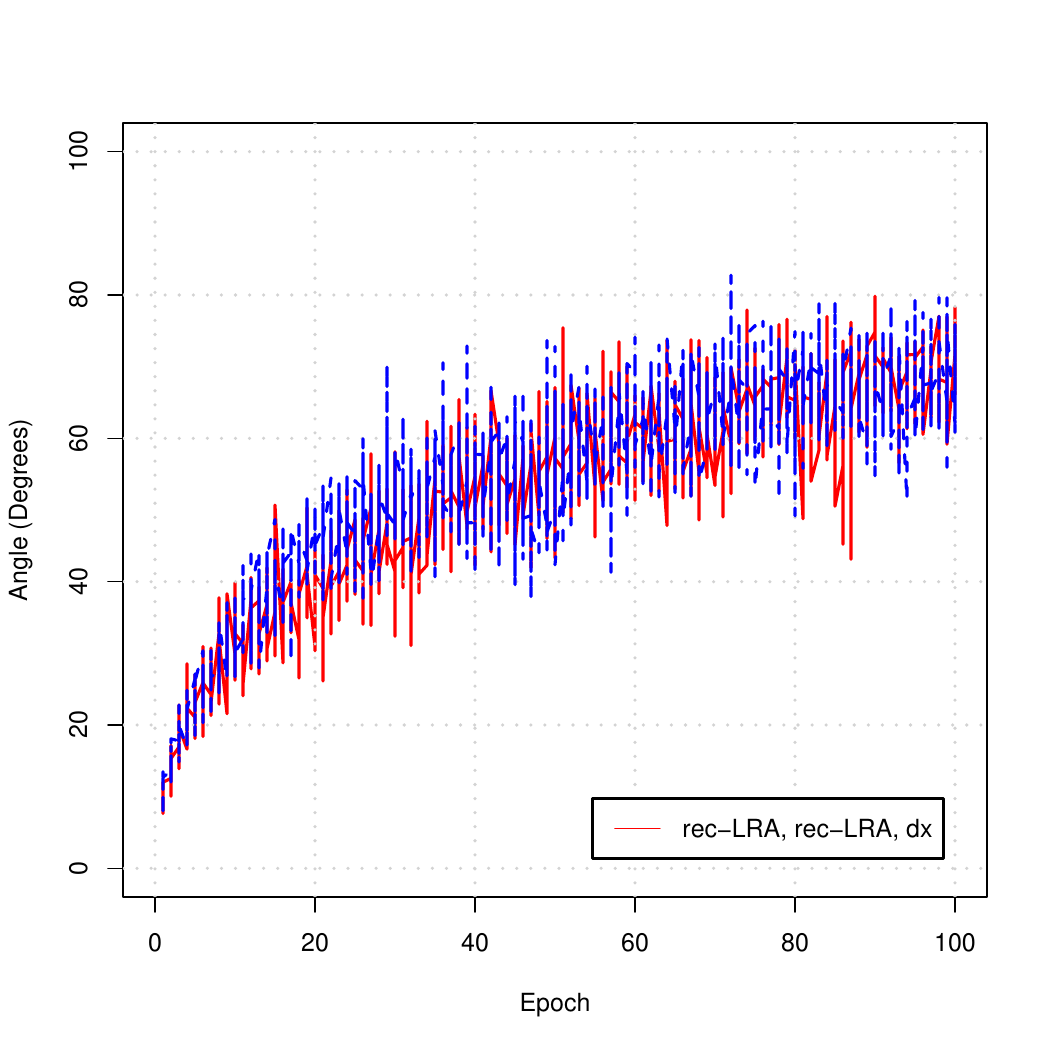}
\vspace{-0.65cm}
\caption{Measured angles between updates given by backprop and: 1) rLRA (red), or 2) rLRA, dx (blue).}
\label{fig:angles}
\vspace{-0.35cm}
\end{center}
\end{wrapfigure}
one of the simplest ways to create targets is to introduce a simple, learnable generative structure that projects errors from one layer down to the one below it \cite{ororbia2018biologically} using another set of parameters called error synapses.
These error weights could also be likened to the decoder weights of the inverse mapping of target propagation \cite{lee2015targetprop}, however, these weights project error messages while target propagation decoder weights project perturbed post-activation activities. One might observe that there are many different possible designs for generative structure of the error synaptic pathways, even beyond the simple skip patterns explored in this paper.
While rec-LRA employs a complex generative structure that entails skipping across long distances to transmit mismatch information from one region to another, one could simply opt for a less intricate pattern, such as a pair-wise transmission structure, where, in the notation of the main paper, the error weight (matrix) connecting any two layers, $(\ell+1)$ and $\ell$, would be denoted $E_{(\ell+1) \rightarrow \ell}$. Note that this is the same as a skip of $g$ = 1 (we analyze the effect of increasing $g$ later in the appendix).
Whatever the generative structure might be for a network with $L$ layers, the target computation for any layer $\ell$ using error (synaptic) cables that originate from some other layer $j \neq \ell$ will always be:
\begin{align*}
    \mathbf{y}_\ell = \phi_\ell(\mathbf{h}_\ell - \beta \mathbf{d}_\ell), \: \mbox{where, } \mathbf{d}_\ell = E_{j \rightarrow \ell} \cdot \mathbf{e}_j
\end{align*}
where we see that the error neurons play a further role beyond their use in updating the forward synaptic weights. Specifically, their information content is first projected down to the layer below (stored as signal $\mathbf{d}_\ell$, which is the displacement vector for latent state $\mathbf{z}_\ell$) and then used to adjust the original pre-activation of $\ell$ through a single weighted integration step.
Much like that in the case of the forward weights, the updates to the error weights themselves also follow from the total discrepancy function:
\begin{align*}
    \Delta E_{j \rightarrow \ell} &= \frac{\partial \mathcal{D}(\Theta)}{\partial E_{j \rightarrow \ell}} = \frac{\partial \mathcal{D}(\Theta)}{\partial \mathbf{y}_\ell} \frac{\partial \mathbf{y}_\ell}{\partial E_{j \rightarrow \ell}} \\
    &= \frac{\partial \mathcal{D}(\Theta)}{\partial \mathbf{y}_\ell} \frac{\partial \phi_\ell(\mathbf{h}_\ell - \beta \mathbf{d}_\ell) }{\partial \mathbf{d}_\ell} \frac{\partial \mathbf{d}_\ell}{\partial E_{j \rightarrow \ell}} \\
    &= \big( -\mathbf{e}_\ell \otimes \phi^{\prime}_\ell(\mathbf{h}_\ell - \beta \mathbf{d}_\ell) \big) \frac{\partial (\mathbf{h}_\ell - \beta \mathbf{d}_\ell) }{\partial \mathbf{d}_\ell} \frac{\partial \mathbf{d}_\ell}{\partial E_{j \rightarrow \ell}} \\
    &= -\beta \big( -\mathbf{e}_\ell \otimes \phi^{\prime}_\ell(\mathbf{h}_\ell - \beta \mathbf{d}_\ell) \big) \cdot (\mathbf{e}_j)^T \approx -\beta \big( -\mathbf{e}_\ell \big) \cdot (\mathbf{e}_j)^T \mbox{.}
    \vspace{-0.2cm}
\end{align*}
While the derive update rule above would work with updating the error synapses (provided that we multiply it by $-\gamma/\beta$), in this work we experimented with two other error update alternatives. One was a simpler Hebbian rule (as was presented in the main paper): $\Delta E_\ell = \gamma(-\mathbf{d}_\ell \cdot (\mathbf{e}_j)^T)$ and the other was: $\Delta E_\ell = \gamma(\mathbf{z}_\ell \cdot (\mathbf{e}_j)^T$. In practice, we have found these two alternative Hebbian rules to yield faster convergence in general -- the first one proved to be useful for the large-scale models (trained on CIFAR-10 and ImageNet) and the second one proved useful for the MNIST/FMNIST experiments. The second rule is similar to \cite{ororbia2018biologically}, prescribing that error weights are adjusted via a product of incoming source error messages they receive and the latent states they wire to.

So long as the updates given by rec-LRA are within $90^{\circ}$ of the gradients given by backprop (which greedily follows the path of steepest descent), the algorithm will move network parameters in the same general direction as backprop and still locate good local optima \cite{nokland2016direct,ororbia2018biologically}. While we defer a formal proof of this algorithmic angle relationship for future work (where one could adapt a proof form of similar structure to that of \cite{nokland2016direct}), we offer empirical support of this fact in Figure \ref{fig:angles}. In this experiment, we measured the angle between two versions of rec-LRA and backprop every $100$ mini-batches throughout the course of a full $100$ epoch training simulation for the $8$-layer residual architecture trained on MNIST in the main paper (but with relu activation functions). The first version of rec-LRA (rLRA) used the error Hebbian rule presented above and the second version (rLRA, dx) utilized the unaltered, derived update rules (which included activation function derivatives). As observed in Figure \ref{fig:angles}, the updates computed by either version of rec-LRA do appear to indeed yield updates are within $90^\circ$ those that would be calculated by backprop, though they appear to be closer to backprop at the start of learning and converge to roughly just under $75^\circ$ and remain relatively stable throughout the learning process. Note that it appears that rec-LRA updates are bit further away from backprop than pair-wise LRA \cite{ororbia2018biologically} (reported at $\sim 40^\circ$).

\begin{figure}[t]
\centering
\includegraphics[width=0.475\textwidth]{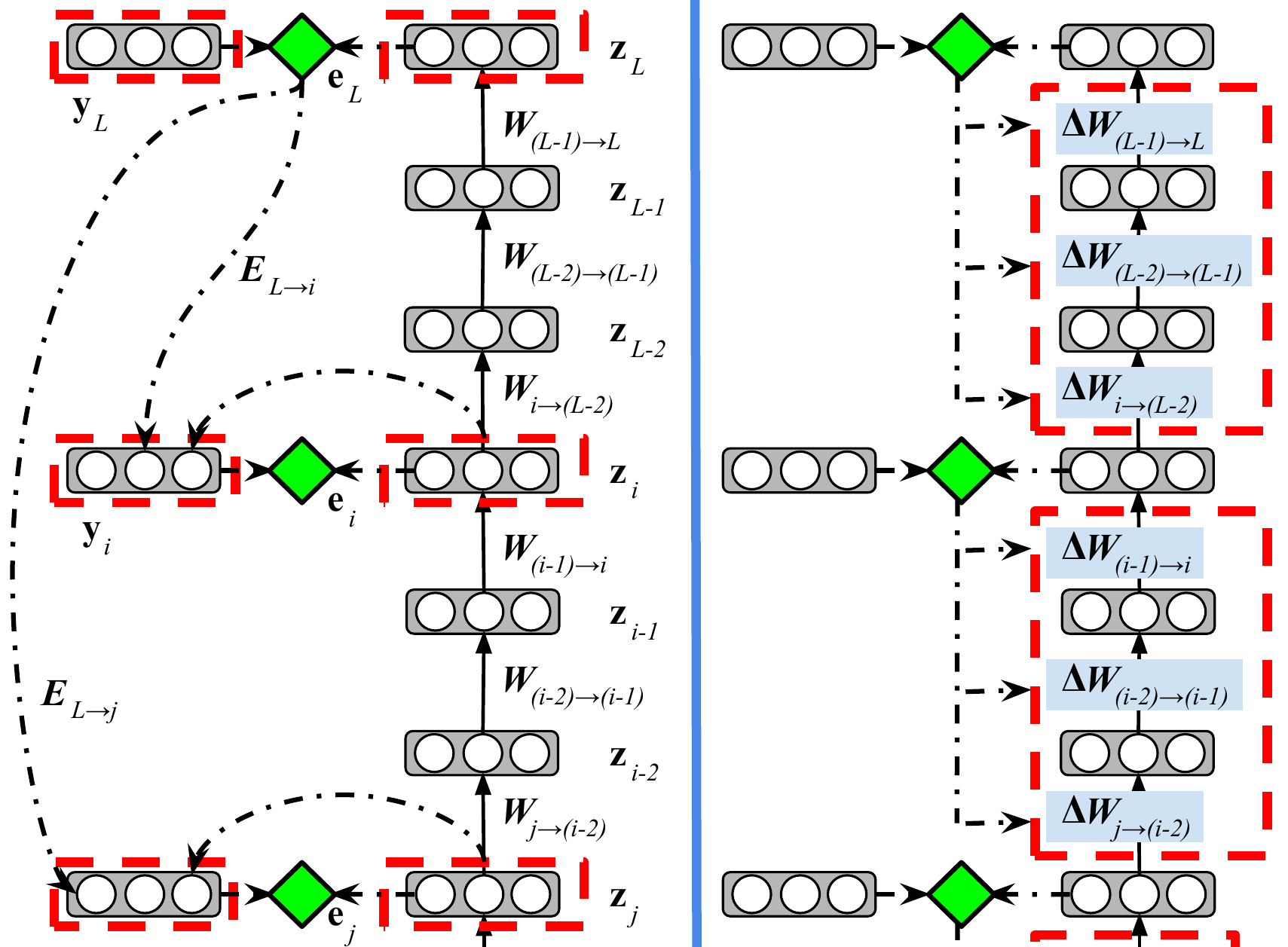} 
\caption{Hybrid rec-LRA -- mixing together LRA target creation (error neurons are green diamonds) with backprop to compute weight updates. Left of blue line depicts target creation \& right depicts (pseudo-)gradient calculation.}
\label{fig:hybrid_reclra}
\vspace{-0.5cm}
\end{figure}

\section*{General Recursive Local Representation Alignment}
\label{sec:general_lra}

The algorithm presented in this section presents rec-LRA in an architecture-agnostic form, which would be suitable for arbitrary operator graphs. Note that, as mentioned in the main paper, rec-LRA can be hybrized with backprop or alternative weight update rules (such as those based on local Hebbian rules). This combination is depicted in Figure \ref{fig:hybrid_reclra}.

\begin{algorithm*}[t]
\caption{The general recursive local representation alignment algorithm.}
\label{algo:rec_lra}
\begin{algorithmic}[1]
    \State {\bf Inputs:} $\mathbf{x}$, $\mathbf{y}$, $f_\Theta(\mathbf{x})$, $\beta$, $\gamma$
    \LineComment{Routine for computing parameter updates for function $f_\Theta(\mathbf{x})$}
    \Function{ComputeUpdates}{$\mathbf{x}$, $\mathbf{y}$, $f_\Theta(\mathbf{x})$}
        \State $\mathcal{H}, \mathcal{Z} = \Call{RunInference}{\mathbf{x}, \mathbf{y}, f_\Theta(\mathbf{x})}$ \Comment Get pre-activities $\mathcal{H}$ \& post-activities $\mathcal{Z}$
        \State $\mathcal{E}, \vartheta = \Call{ComputeErrorNeurons}{\mathbf{y}, f_\Theta(\mathbf{x}), \mathcal{H}, \mathcal{Z}}$, \quad $\Delta_{\text{all}} = \emptyset$
        \For{$W_{i \rightarrow j} \in \Theta$}
            \State $\mathbf{z}_i \leftarrow \mathcal{Z}[i]$, \quad 
            $\mathbf{e}_j \leftarrow \mathcal{E}[j] $ 
            \State $\Delta W_{i \rightarrow j} = \mathbf{e}_j \cdot (\mathbf{z}_i)^T $, \quad $\Delta_{\text{all}} = \Delta_{\text{all}} \cup \{\Delta W_{i \rightarrow j}\}$
        \EndFor
        \For{$E_{j \rightarrow i} \in \Theta$}
            \State $\mathbf{d}_i \leftarrow \vartheta[i]$, 
            \quad $\mathbf{d}_j \leftarrow  \vartheta[j]$, 
            \State $\Delta E_{j \rightarrow i} = -\gamma(\mathbf{e}_i \cdot (\mathbf{d}_j)^T) $ \quad $\Delta_{\text{all}} = \Delta_{\text{all}} \cup \{\Delta E_{j \rightarrow i}\}$
        \EndFor
        \State \Return $\Delta_{\text{all}}$ \Comment Return full set of parameter updates to $\Theta$
    \EndFunction
    
   \LineComment{ Routine for calculating all error neuron \& delta signal vectors for function $f_\Theta(\mathbf{x})$}
   \Function{ComputeErrorNeurons}{$\mathbf{y}_\ell$, $f_\Theta(\mathbf{x})$, $\mathcal{H}$, $\mathcal{Z}$ } 
        \State $\mathcal{E} = \{ \emptyset \} * L$, $\vartheta = \{ \emptyset \} * L$ \Comment Initialize arrays w/ empty error neurons \& delta signals
        \State $\Call{ComputeSignals}{L, \mathbf{y}_\ell, f_\Theta(\mathbf{x}), \mathcal{H}, \mathcal{Z}, \mathcal{E}, \vartheta}$
        \State \Return $\mathcal{E}$, $\vartheta$
   \EndFunction
   
   \LineComment{Sub-routine meant to support the routine $\Call{ComputeErrorNeurons}{\circ}$}
   \Function{ComputeSignals}{$\ell$, $\mathbf{y}_\ell$, $f_\Theta(\mathbf{x})$, $\mathcal{H}$, $\mathcal{Z}$, $\mathcal{E}$, $\vartheta$ } 
        \State $\mathbf{z}_\ell = \mathcal{Z}[\ell]$, \quad $\mathbf{e}_\ell = (\mathbf{z}_\ell - \mathbf{y}_\ell)$, \quad $\mathcal{E}[\ell] \leftarrow \mathbf{e}_\ell$
        \State $\Upsilon \leftarrow \Call{ExtractChildrenIndices}{\ell, f_\Theta(\mathbf{x})}$ \Comment Get children node indices for error node $\ell$
        \LineComment{Recursive Case: Traverse into each children error node \& compute its error vector}
        \If{$\Upsilon$ is not $\emptyset$}
        \For{$i \in \Upsilon$}
            \State $\mathbf{h}_i = \mathcal{H}[i]$,\quad  $\mathbf{d}_i = E_{\ell \rightarrow i} \cdot \mathbf{e}_\ell$,\quad  $\mathbf{y}_i = \phi_i(\mathbf{h}_i - \beta \mathbf{d}_i)$, \quad $\vartheta[i] \leftarrow \mathbf{d}_i$
            \State $\Call{ComputeSignals}{i, \mathbf{y}_i, \mathcal{G}_\Theta, f_\Theta(\mathbf{x}), \mathcal{H}, \mathcal{Z}, \mathcal{E}, \vartheta }$
        \EndFor
        \EndIf
   \LineComment{Base Case: No children error neurons, so no need to further update graph at node $\ell$}
   \EndFunction
\end{algorithmic} 
\end{algorithm*}

After running the architecture's forward pass procedure to gather layer-wise activities, rec-LRA computes mismatch signals by starting at the layer $L$ and computing the corresponding error neurons $\mathbf{e}_L$. From there, rec-LRA retrieves the layer indices of the regions that immediately connect to $L$ (via an implementation of the function $\Call{ExtractChildrenIndices}{\circ}$), storing these in the array $\Upsilon$.  $\Upsilon$ is an un-ordered list of integers, since transmitting the mismatch signal from $j$ to $i$ does not depend on the transmission from $j$ to $\Upsilon \setminus i$. This means that the transmission of mismatch signals to each of $L$'s neighbors can be done in parallel if multiple processors are available. 
For a target region $i$ connected to $L$, rec-LRA will compute its target $\mathbf{y}_i $.
It will then recursively call itself on that region using the newly computed target, subsequently computing the error neuron vector at $i$ and further computing targets for any regions $\Upsilon$ connected to $i$ and so on and so forth. The base case for termination in full rec-LRA is simply the situation when $\Upsilon = \emptyset$, i.e., there are no regions that immediately connect to $i$. Once all error neuron vectors have been computed, we can calculate updates to all parameters of not only the neural architecture but also to each error matrix used to transmit $\gamma$ is a decay factor (typically set close to $1.0$) meant to ensure that the error weights change more slowly than the forward weights.
While the pseudocode in Algorithm \ref{algo:rec_lra} first computes the error neurons ($\Call{ComputeErrorNeurons}{\circ}$) then calculates parameters updates (lines 6-11 in $\Call{ComputeUpdates}{\circ}$) after, one could actually merge the two functions together and immediately compute the updates for any incoming model weights $W_{k \rightarrow i}$ that connect to region $i$ as well as relevant error weights $E_{j \rightarrow i}$. Furthermore, even though the algorithm as presented would execute each recursive call sequentially (in the sub-routine $\Call{ComputeSignals}{\circ}$), given that transmission of error from $j$ to $i$ is independent of that from $j$ to $\Upsilon \setminus i$, one could allocate each call to a cluster/set of CPUs/GPUs dedicated to generating targets for the parts of the operator graph that the call will see.
This design highlights one of rec-LRA's key strengths -- it compute targets and parameter updates in a divide-and-conquer approach using pathways defined by error connectivity.

\section*{Experimental Details and Further Analysis}
For the parameter optimization carried out on the MNIST and Fashion MNIST benchmarks, we employed the Adam \cite{kingma2014adam} adaptive learning rate, using a learning rate of $\lambda = 2e-4$ (tuned using validation performance for each dataset). Updates to parameters, whether they were calculated via backprop or rec-LRA, were estimated over mini-batches of size $32$ and layers (both forward and error synaptic weights) were initialized according to an element-wise, zero-mean Gaussian distribution with standard deviation $\sigma$ selected in the range of $[0.025-0.1]$ (tuned using validation performance). Models with $5$ layers of $256$ units were trained over $500$ epochs and those with $8$ layers of $256$ neurons were trained for $100$ epochs.
Note that we chose this configuration (including \# of epochs) to be comparable to related prior work \cite{bartunov2018assessing}.

\begin{figure}[ht]
\centering
\includegraphics[width=0.4\columnwidth]{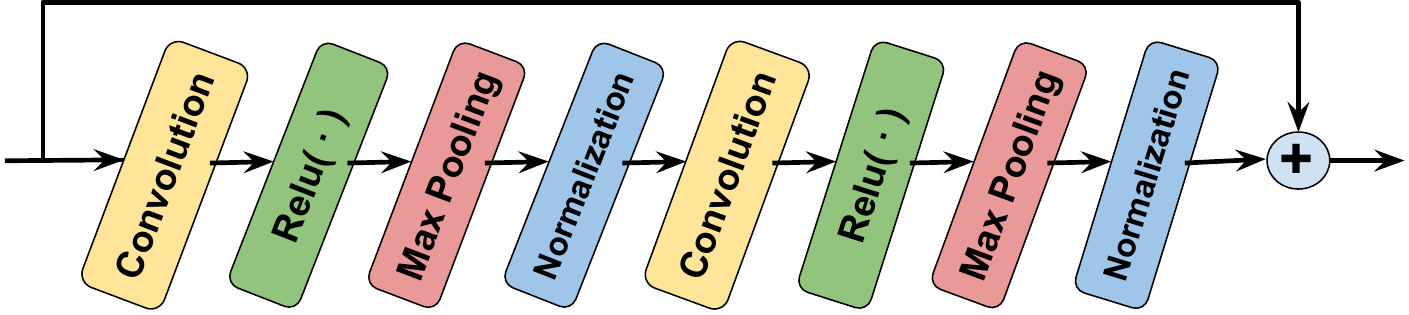}
\caption{The residual block experimented with in this study.}.
\label{fig:res_block}
\vspace{-0.5cm}
\end{figure}

For the CIFAR-10 and ImageNet benchmarks, with respect to image pre-processing, global contrast normalization was applied, where each color channel's pixel mean was subtracted from itself. ZCA whitening was then applied, where: 1) the image data was centered and rotated onto its principle components,  2) the principle components were normalized, and 3) the image was finally rotated back.
Parameter updates were estimated with mini-batches of $10$ samples. To optimize network weights using either rec-LRA or backprop, we employed AdamW \cite{loshchilov2017decoupled} with a global learning rate of $\lambda = 1e-4$ and used both layer normalization and batch normalization in the architectures for regularization.
For the CNN and residual network models, a further generalization of the error neurons was employed. Inspired by the success of combining L1 and L2 losses (similar to elastic net regression) in the domain of neural image compression \cite{ororbia2019learned,mali2020sibling}, we used a convex combination of two sets of error neurons:
\begin{align*}
    \mathbf{e}_\ell = \alpha_e (\mathbf{z}_\ell - \mathbf{y}_\ell) + (1 - \alpha_e) sign(\mathbf{z}_\ell - \mathbf{y}_\ell)
\end{align*}
where $\alpha_e$ is a scalar factor meant to control the trade-off between the two types of error neurons. We found, after preliminary experimentation, that $\alpha_e = 0.19$ for error neurons that exist at the end of skip-error connections (recursive depth $1$, or error neurons embedded at the output of a block of operations) and $\alpha_e = 0.24$ for neurons that exist at the end of error synapses that connect a pair of layers locally (recursive depth $2$, or error neurons embedded within a block of operations).
We used additive noise set to a level of $0.1$, $256$ perturbation masks per layer for ImagNet, and $160$ perturbation masks per layer for CIFAR-10. We trained each model for $100$ epochs and tuned individual meta-parameters based on validation performance. Rec-LRA specific meta-parameters found from validation tuning were found to be $\beta = 0.1205$ and $\gamma = 0.1524$. Note that the residual block we used is depicted in Figure \ref{fig:res_block}.

\begin{wrapfigure}{r}{2.75in}
\begin{center}
\vspace{-0.475cm}
    \includegraphics[width=0.36\textwidth]{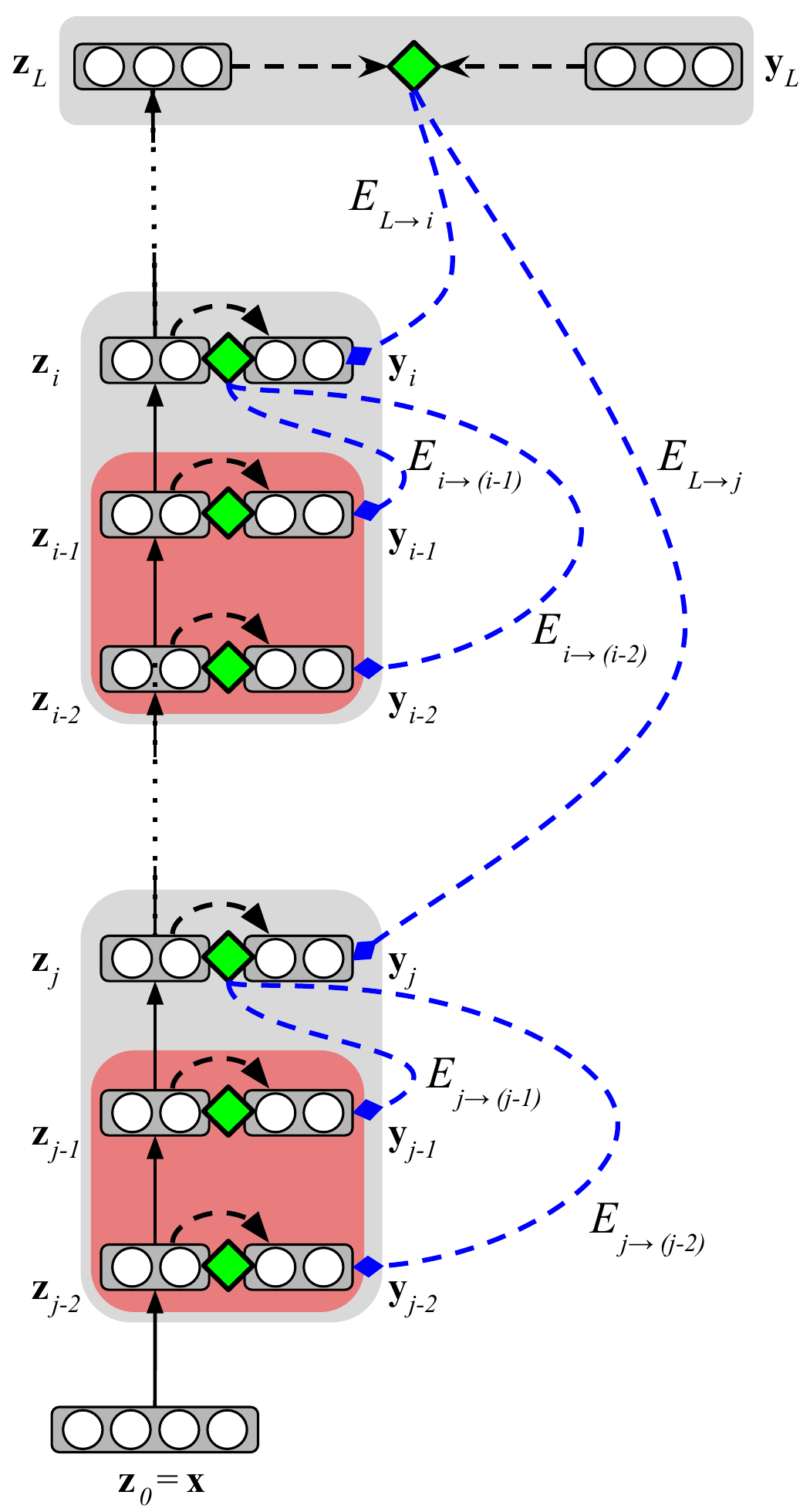}
    \caption{A depiction of recursive-LRA on a feedforward neural, with recursive depth of $2$ shown. Green diamonds represent error neurons while blue dashed lines represent error weights that transmit mismatch signals to specific layers in the network.}
    \label{fig:rec_lra_ann}
\end{center}
\vspace{-0.15cm}
\end{wrapfigure}

The rest of our configuration settings were set to be similar to \cite{juefei2018perturbative} to ensure a fair comparison among models, i.e. we use rec-LRA and BP to train ResNet-18 models \cite{he2016deep}. Layers (forward and error weights) were initialized with a unit Gaussian distribution (Xavier and orthogonal initialization schemes were found to yield unsatisfactory performance).
The architectures of our trained CNN models were set to be identical to those of \cite{bartunov2018assessing}, hence meaning we trained using the locally-connected structure originally proposed in that study as a more biologically-plausible replacement of standard convolution. The update rule for locally-connected receptive field structures are the same as that used for fully-connected weights.

In Figure \ref{fig:rec_lra_ann}, we present another visual of what the error synaptic structure would look for two arbitrary blocks of operations (note, as in the main paper, a block could be a residual operator block or any collection of operators that are designated as belonging to a group). The blue-dashed lines depict the flow of information of error messages from one layer to another, but it should be noted that they visually abstract away the actual, fully-connected error weight matrices that connect any two layers. For example, take $E_{L \rightarrow i}$ -- its concrete instantiation would be a matrix of $|\mathbf{y}_L| \times |\mathbf{z}_i|$, where $|\mathbf{v}|$ measures the dimesionality of vector $\mathbf{v}$ and the green diamond at layer $L$ would be implemented as a vector $|\mathbf{y}_L|$ neurons, since there would be one error neuron per standard neuron in order to measure its mismatch from a corresponding target value. The black solid arrows would be implemented as feedforward weights in the diagram example. The dashed black curved arrows simply imply that an internal target (to the right of a green diamond) would simply be a function of the original latent state (to the left of the green diamond) and weighted displacement signal (transmitted by error weights).

\paragraph{Update Re-Projection / Gradient Re-Scaling: }
For all architectures and algorithms, in all experiments of this paper, we re-projected weight updates (or gradients) back to a Gaussian ball of radius $c$ (as in \cite{pascanu2013difficulty}).
Formally:
\begin{align*}
Nm(\Delta, c) = \Bigg \{\frac{c}{||\Delta||}\Delta \mbox{, if } ||\Delta|| \geq c \mbox{, and }
\Delta \mbox{, if } ||\Delta|| < c \Bigg \} 
\end{align*}
where $\Delta$ is any parameter update matrix returned by a learning algorithm. We found that gradient re-projection was useful to consistently ensure stable training.

\begin{table}[t]
    \caption{(Left) Expanded ImageNet results. (Middle) Generalization performance analysis of networks trained on CIFAR-10 and ImageNet. (Right) CIFAR-10 skip-gap analysis (measuring validation error as a function of gap $g$).}
     \label{results:expanded_results}
     \begin{minipage}{2.35in}
     \vspace{0.1cm}
     \begin{tabular}{l | c c} 
     \hline
     \textbf{ImageNet} & Top-1 & Top-5 \\ 
     \hline
     TP \cite{bartunov2018assessing}& $98.34$ & $94.56$ \\
     DTP-Alternating \cite{bartunov2018assessing} & $99.36$ & $97.28$ \\
     SDTP-Parallel \cite{bartunov2018assessing} & $99.28$ & $97.15$ \\
     FA \cite{bartunov2018assessing} & $93.08$ & $82.54$ \\
     ResNt, FA \cite{xiao2018biologically} & $90.52$ & $77.32$ \\
     ResNt, FA+BP \cite{xiao2018biologically} & $73.01$ & $51.24$ \\
     ResNet, BP \cite{xiao2018biologically} & $33.14$ & $12.49$ \\
     ResNet, SS \cite{xiao2018biologically} & $37.91$ & $16.18$ \\
     ResNet, SS+BP \cite{xiao2018biologically} & $37.01$ & $15.44$ \\
     ResNet, KP \cite{akrout2019weighttransport} & $\mathbf{29.2}$ & N/A \\
     \hline
     CNN, BP & $62.58$ & $39.89$ \\ 
     CNN, rLRA & $73.69$ & $49.78$ \\ 
     ResNet, BP & $\mathbf{28.15}$ & $\mathbf{9.81}$  \\
     ResNet, rLRA  & $\mathbf{30.48}$ & $\mathbf{11.97}$ \\
     \hline
    \end{tabular}
    \end{minipage}%
     \begin{minipage}{2.55in}
     \begin{tabular}{l | c c c c} 
     \hline
     \multicolumn{4}{c}{\textbf{Cifar-10}}  \\
     \hline
     \textbf{Algo.} & Acc & F1 & Prec & Rec \\ 
     \hline
     BP & $91.15$\% & $90.24$ & $89.51$ & $90.58$ \\
     rLRA   & $90.88$\% & $89.96$ & $89.20$ & $88.61$ \\
     \hline 
      \multicolumn{4}{c}{}  \\
     \hline
     \multicolumn{4}{c}{\textbf{ImageNet}}  \\
     \hline
     \textbf{Algo.} & Acc & F1 & Prec & Rec \\ 
     \hline
     BP & $71.84$\% & $70.04$ & $70.12$ & $69.96$ \\
     rLRA  & $69.52$\% & $67.82$ & $68.12$ & $67.52$ \\
     \hline
    \end{tabular} 
    \end{minipage}%
    \begin{minipage}{1.4in}
    \vspace{-2.7cm}
      \begin{tabular}{l | c c } 
      \hline
      \multicolumn{2}{c}{\textbf{CIFAR-10}} \\
      \hline
      Gap $g$ & Valid Error  \\
      \hline
      Skip-$1$  & $13.98$ \\
      skip-$2$  & $6.12$  \\
      skip-$3$ & $9.58$    \\
      skip-$4$ & $11.28$  \\
      skip-$5$ & $12.58$ \\
      skip-$6$ & $10.08$ \\
     \hline
     \end{tabular}
     \end{minipage}%
\end{table}

\paragraph{Expanded Results, Analysis, \& Latent Visualization:} We present in Table \ref{results:expanded_results} (Left) an expanded table of results for ImageNet that include some additional relevant algorithm measurements. We report the performance of the (best-performing) sign symmetry (SS) of \cite{xiao2018biologically}, which we outperform though the margin of improvement is far narrower. 
It is important to note that the best version of SS we report still utilizes partial backprop in its calculations while rec-LRA is gradient-free.
In Table \ref{results:expanded_results} (Middle), for both CIFAR-10 and ImageNet, we dissect the networks' predictions (beyond accuracy) by analyzing their confusion matrices on the test set -- we calculate precision (Prec), recall (Rec), and the F1 score (the harmonic mean between Prec and Rec). For CIFAR-10, in terms of these metrics, it appears that rec-LRA is a bit weaker in recall comapred to backprop, though its precision is quite close to that of backprop. 
We speculate that the small gap in performance could be closed with a more rigorous tuning of the meta-parameters of rec-LRA on the validation set. Nonetheless, rec-LRA's strong generalization on CIFAR-10 already offers evidence of its ability to scale up to a more challenging problem involving color images.

On ImageNet (Table \ref{results:expanded_results} (Middle)), we observe a bit of a larger performance gap, especially in terms of recall. However, with only minimal tuning of the ResNet trained via rec-LRA, its generalization performance is quite impressive. We hypothesize that with more rigorous/careful tuning and the use of additional model heuristics, performance will improve across all metrics.
In the last table, i.e., Table \ref{results:expanded_results} (Right), we measure the validation error of the CNN model's chosen skip gap $g$ (for recursive depth $2$) when using rec-LRA (since a CNN does not have a natural residual block structure to exploit when crafting the error structure for rec-LRA, skip could be treated as a meta-parameter). Observe that all skip gaps $g > 1$ yield a reduction in error over a simple layer pair-wise error structure ($g = 1$) but there appears the best error for CIFAR-10 is with a skip of $g = 2$ (though it is possible that gaps beyond $6$, or, better, using recursive depths greater than $2$, might improve generalization).

\begin{figure}[t]
\centering
  \includegraphics[width=.35\linewidth]{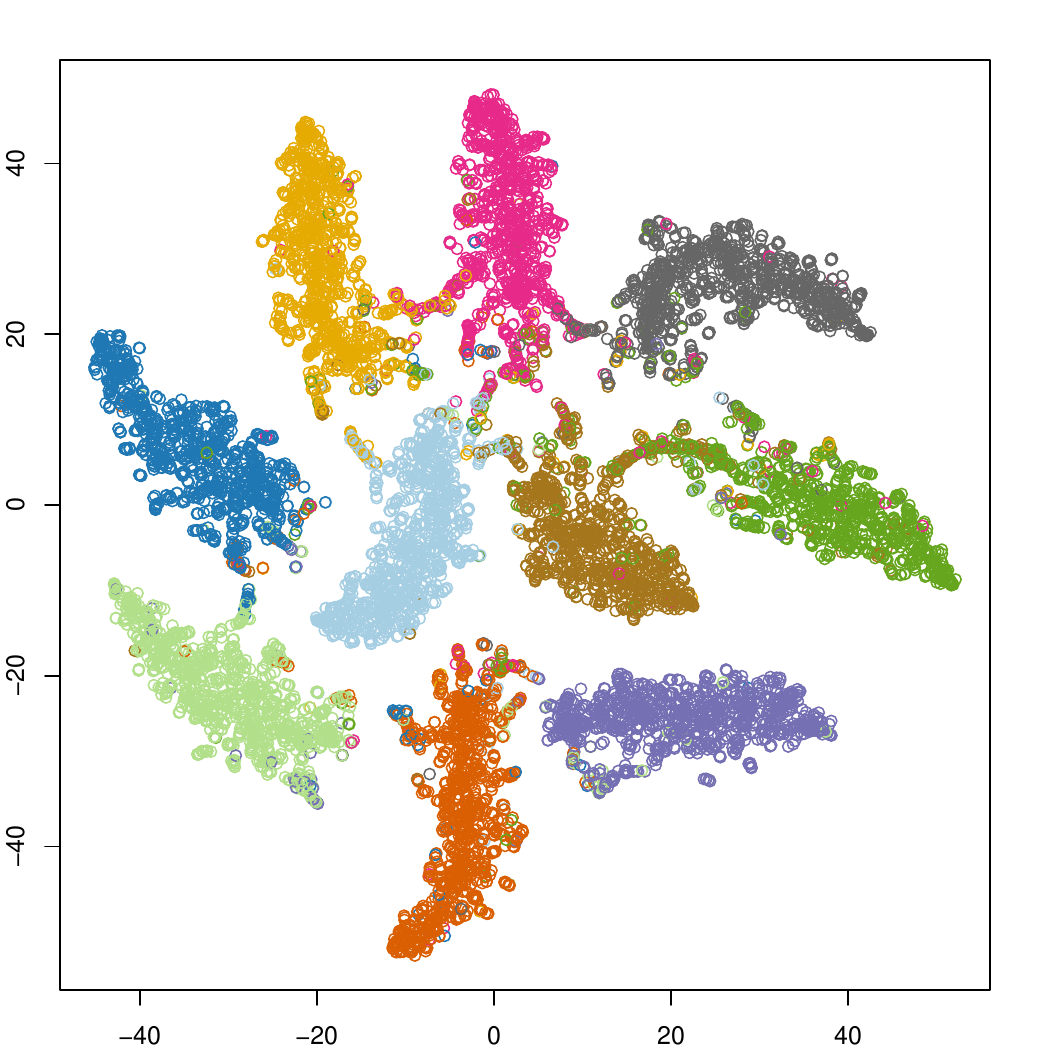}
  ~
  \includegraphics[width=.35\linewidth]{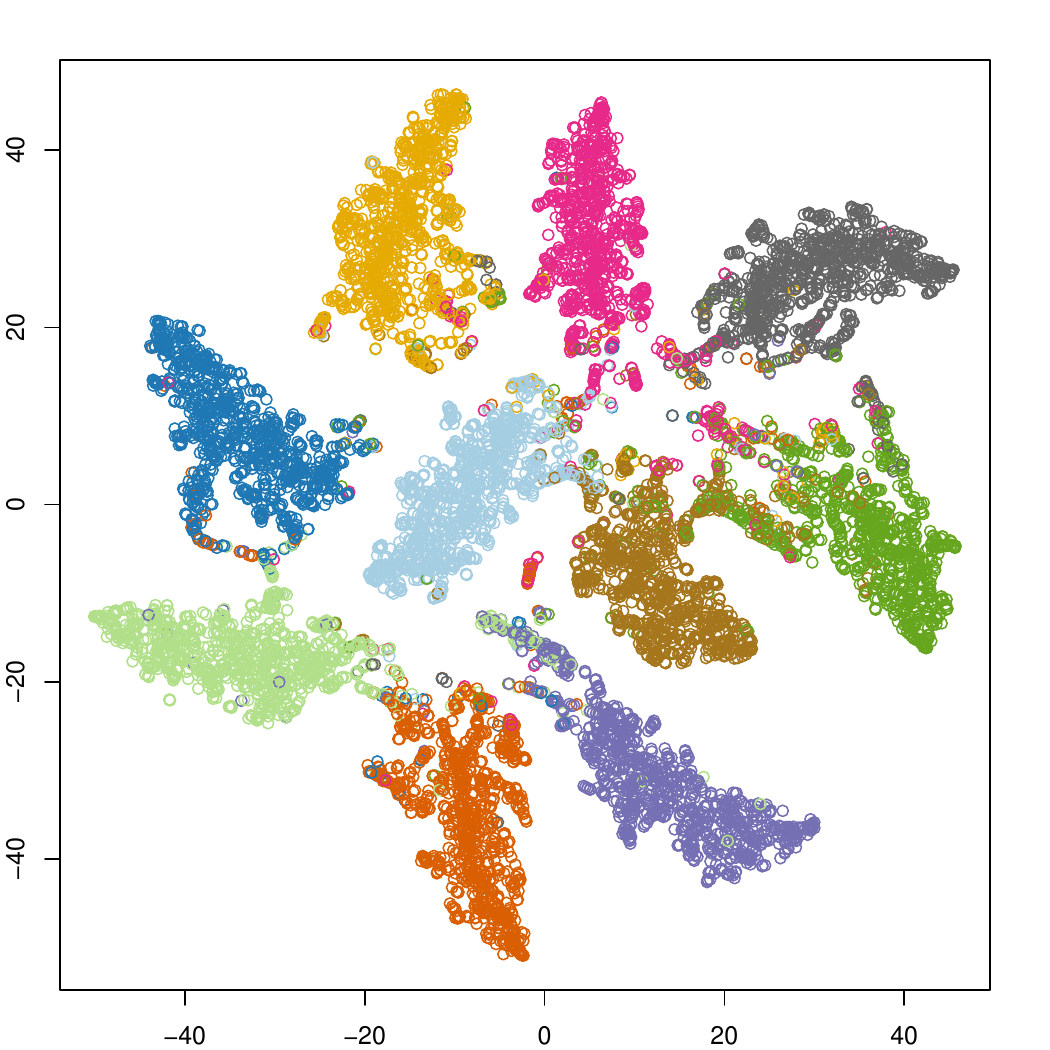}
  \caption{t-SNE visualization of Resnet trained with either backprop (left) or rec-LRA (right).}
  \label{fig:resnet_tsne}
\end{figure}

Finally, in Figure \ref{fig:resnet_tsne}, for the trained CIFAR-10 networks, we visualize the top-most latent representations acquired by those trained by backprop and rec-LRA, using t-SNE \cite{van2013barnes}. Perplexity was set to $30$ and $100$ iterations were used to fit t-SNE on the latents. Qualitatively, we observe that rec-LRA does indeed learn a good separation/clustering of classes in its latent representations (just as backprop does).

\end{document}